%% file: access.tex
\documentclass{ieeeaccess} % this class wraps IEEEtran

\usepackage{cite}
\usepackage{amsmath,amssymb,amsfonts}
\usepackage{algorithmic}
\usepackage{graphicx}
\usepackage{textcomp}
\usepackage{url}
\usepackage{listings}
\usepackage{booktabs}
\usepackage{float} % for the [H] placement specifier
\usepackage[dvipsnames]{xcolor}
\usepackage{dsfont}
\usepackage{multirow}
\usepackage{xcolor}
\usepackage{hyperref}

\definecolor{navyblue}{rgb}{0.0, 0.0, 0.5}

\hypersetup{
    colorlinks=true,
    linkcolor=black,        % Internal document links (e.g., Fig. 1, Eq. 2)
    citecolor=black,        % Bibliography citation numbers
    urlcolor=accessblue           % External web/GitHub links
}

\def\BibTeX{{\rm B\kern-.05em{\sc i\kern-.025em b}\kern-.08em
  T\kern-.1667em\lower.7ex\hbox{E}\kern-.125emX}}

\usepackage[utf8]{inputenc}
\usepackage{xcolor}
\usepackage{listings}

\definecolor{olivegreen}{rgb}{0,0.6,0}
\definecolor{purple}{rgb}{0.5,0,0.5}
\definecolor{navyblue}{rgb}{0,0,0.5}

\lstdefinelanguage{turtle}
{
    morekeywords={PREFIX,a},
    morecomment=[l]{\#},
    morestring=[b]",
    sensitive=true
}

\definecolor{olivegreen}{rgb}{0,0.6,0}
\definecolor{purple}{rgb}{0.5,0,0.5}
\definecolor{highlight}{rgb}{0.952, 0.447, 0.172} 

\begin{document}

% --------------------------------------------------
% FRONT MATTER METADATA
% --------------------------------------------------
\history{Received August 8, 2026.}
\doi{10.1109/ACCESS.2026.DOI}

% \title{H2:Introducing a hybrid rule-based and LLM-driven framework for constructing medical knowledge graph from raw metadata}

% \title{H2:A hybrid document, schema-on-the-read data lake with a hybrid rule-based and LLM-driven mechanism for constructing medical knowledge graph from raw metadata}

% \title{$H_2$: A Dual Hybrid Document-based and Schema-on-Read Architecture with Rule-based and LLM-driven Mechanisms for Medical Knowledge Graph Construction}
\title{$H_2$: A Dual Hybrid Semantic Data Lake Architecture for Medical Data Harmonization with Human-In-the-Loop verified, LLM Driven Metadata Annotation System }

\author{
  \uppercase{Ioannis N. Tzortzis}\authorrefmark{1}, \uppercase{Georgia Kapetadimitri}\authorrefmark{2}, 
  \uppercase{Agapi Davradou}\authorrefmark{1}, 
  \uppercase{Nefeli Kousta}\authorrefmark{1}, 
  \uppercase{Nikolaos Bakalos}\authorrefmark{1}, 
  \uppercase{Ioannis Rallis}\authorrefmark{1}, 
  \uppercase{Dimitrios Kalogeras}\authorrefmark{1}, 
  \uppercase{Nikolaos Doulamis}\authorrefmark{1}, and 
  \uppercase{Anastasios Doulamis}\authorrefmark{1}\\
  % \thanks{All authors are with the Institute of Communication and Computer Systems (ICCS),  15780 Zografou, Athens, Greece (e-mail: i.n.tzortzis@gmail.com).}
}

\address[1]{Institute of Communication and Computer Systems (ICCS),  15780 Zografou, Athens, Greece.}

\address[2]{University of Macedonia (UOM),  54636, Thessaloniki, Greece.}

\markboth
{Tzortzis \headeretal: $H_2$: A Dual Hybrid semantic data lake
architecture for medical data harmonization}
{Tzortzis \headeretal: $H_2$: A Dual Hybrid semantic data lake
architecture for medical data harmonization}

\corresp{Corresponding author: Ioannis N. Tzortzis (itzortzis@mail.ntua.gr).}

% --------------------------------------------------
% ABSTRACT AND KEYWORDS
% (for THIS ieeeaccess.cls, these MUST come BEFORE \maketitle;
%  the class stores them in boxes and prints them inside \@maketitle)
% --------------------------------------------------
\begin{abstract}
Medical data, by its nature, exhibit a high degree of heterogeneity on multiple levels ranging from (a) different modalities like images, text and time series, (b) diverse  tabular schemata introduced by institutions and (c) completely unstructured textual information data provided by healthcare professionals. Data lakes are often used in medical data storage to consolidate all heterogeneous diverse data in a single, central location, where it can be saved "as is", without the need to impose a schema like a data warehouse does.  Despite their flexibility, though, data lakes are notorious for the “data swamp” failure. Thus, providing a reliable data harmonization mechanism through metadata, without compromising integrity or flexibility, is a real challenge. To this end, knowledge graphs have attracted attention since they provide a dynamic way to depict relationships without a rigid schema-on-write approach. Additionally, another rigorous task relies on the interoperability of data: application of appropriate ML techniques on such a diverse nature of data is not an easy task, as a domain expert must decide the efficacy of a method to a specific data type or dataset. Metadata annotation can aid by tagging applicable operations, however this requires manual intervention, not to mention the plethora of existing datasets which lack such information. To tackle both challenges, in this paper, we propose a semantic data lake architecture that promotes data harmonization and incorporates a generative annotation process (i.e. LLMs) of non-labeled metadata collections to support the application of meaningful ML techniques. Building on top of this approach, we create a higher level of knowledge, identifying suitability of data with respect to applicable ML operations based on their data nature. We utilize a constrained generation approach with specific rules to avoid hallucinations and ensure that produced outputs are limited by the available ML techniques. To prove the validity of our approach, we tested this method with metadata collection from datasets in Kaggle which were enriched with tags identified in the constrained superset of ML techniques. As those tags were removed from the dataset metadata, we prompted (i.e. asked) the LLM to generate them in order to test the applicability of the proposed operation. This constrained metadata generation problem appears as a multi-label classification task thus a recall score is calculated. We examined various LLM vs the recall metric and generation speed to assess how model sizes influence the applicability of LLMs for an accurate metadata annotation. Additionally, we propose a Human-in-the-Loop approach to ensure metadata integrity and consistency by allowing the operator to validate the LLM generated tags. Based on the results, we selected Gemma 3: 4B to be deployed in a data lake operation, as it presented the best combination of recall,  speed across various  datasets while maintaining low resource requirements.
% This paper introduces a backend architecture for modern data lakes, leveraging a flexible design to ensure adaptability and scalability. While a document-oriented database is employed to store raw metadata, a schema-on-read approach is adopted to generate semantic knowledge for the construction of a Knowledge Graph. Core Resource Description Framework (RDF) triples are generated through rule-based processing of the raw metadata, while additional entries are created by a Large Language Model (LLM) acting as an AI specialist in data categorization. To fully exploit the flexibility of this hybrid system and the enhancements provided by the LLM, we introduce a Human-in-the-Loop approach at the data ingestion point to ensure metadata integrity and consistency. In this work, we experiment with several state-of-the-art LLMs from three different families, categorized into three tiers (low, medium, high) according to their parameter cardinality, to validate our concept and investigate the performance of available solutions. According to the results, the $Gemma 3:4B$ is the superior model in cases of $D_1$ and $D_2$ while the $Gemma3:27b$ model outperforms the rest models in $D_3$ case, with Recall reaching as high as 78.8\%, 80.1\% and 68.3\% respectively.
\end{abstract}

\tfootnote{This work has been submitted to the IEEE for possible publication. Copyright may be transferred without notice, after which this version may no longer be accessible.}


\maketitle

\input{sections/introduction}

\input{sections/related_work}

\input{sections/architecture}

\input{sections/methodology}

% \subsection{Dynamic Knowledge Graph Generation}

\input{sections/results}

% \newpage
\section{Conclusion}
\label{sec:conclusion}
\input{sections/conclusion}

\clearpage
\bibliographystyle{IEEEtran}
\bibliography{references}

\vspace{5cm}
% \raggedbottom
\begin{IEEEbiography}
[{\includegraphics[width=1in,height=1.25in,clip,keepaspectratio]{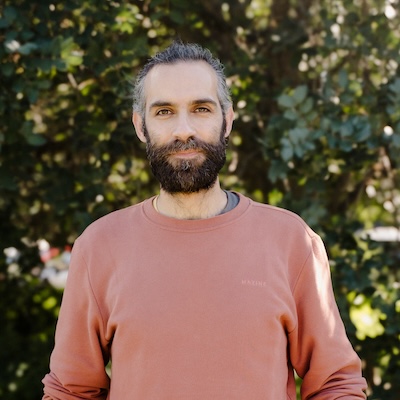}}] {Ioannis N. Tzortzis} received his Diploma in Electronics and Computer Engineering from the Technical University of Crete (TUC) in 2016, with a thesis focusing on the development of a compact regenerative braking system for electric vehicles. He later earned an M.Sc. in Automation Systems from the National Technical University of Athens (NTUA), where his research focused on developing a yaw rate-based control system to adjust the camber angle of the front wheels on a prototype vehicle. Currently, he is a Ph.D. candidate investigating the application of machine learning and computer vision techniques in medical data and healthcare applications in general. 
\end{IEEEbiography}

\begin{IEEEbiography}
[{\includegraphics[width=1in,height=1.25in,clip,keepaspectratio]{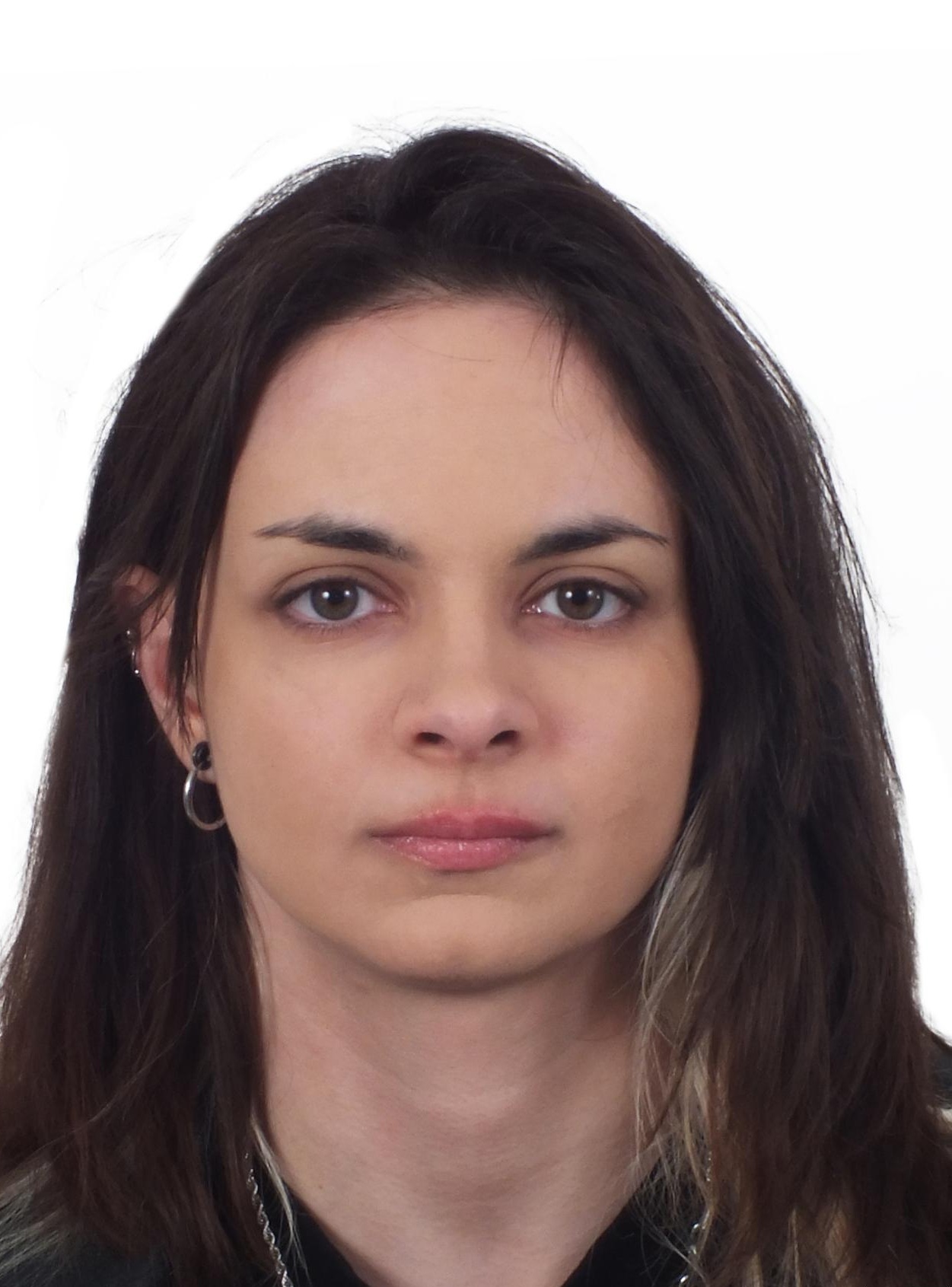}}]{Georgia Kapetadimitri}  is a Junior Researcher at the Laboratory of Photogrammetry, National Technical University of Athens (NTUA). She is also pursuing a PhD in the Department of Applied Informatics at the University of Macedonia. Her research interests include explainable AI, graph-based deep learning, deep learning for remote sensing and photogrammetry, and applications of machine learning to Earth observation data.
\end{IEEEbiography}

\begin{IEEEbiography}
[{\includegraphics[width=1in,height=1.25in,clip,keepaspectratio]{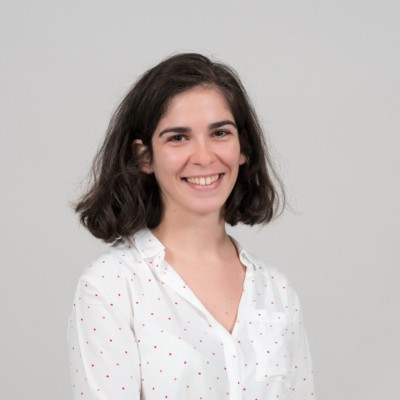}}]{Agapi Davradou} received the integrated M.Sc. degree in production engineering and management from the Technical University of Crete, Chania, Greece, in 2018. She received the the M.Sc. degree in advanced informatics and computing systems, specializing in software development and artificial intelligence, from the University of Piraeus, Athens, Greece, in 2021. This degree was in collaboration with the Instituto Superior Técnico, Lisbon, Portugal, where she completed her master's thesis focusing on detection and segmentation of pancreas in CT scans.
Currently, she is a Senior Data Scientist at Dialectica. In the past, she has collaborated as a part-time researcher with the National Technical University of Athens (NTUA). Her research interests focuses on the application of deep learning to medical data and healthcare applications.
\end{IEEEbiography}

\begin{IEEEbiography}
[{\includegraphics[width=1in,height=1.25in,clip,keepaspectratio]{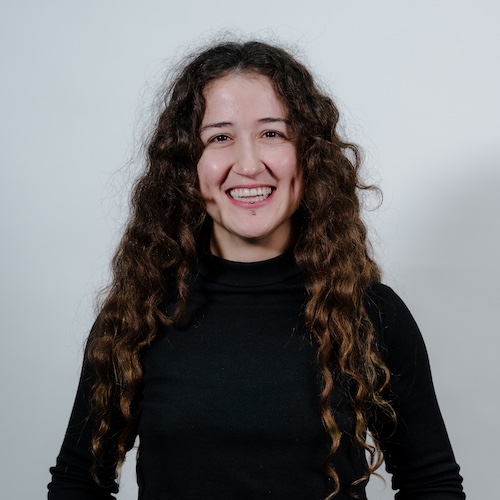}}]{Nefeli Kousta} is currently a research assistant with ICCS. She holds an MSc in Data Science and Machine Learning from National Technical University of Athens, where she also earned an Integrated Master’s degree in Applied Mathematical and Physical Sciences. Her research interests span data science, deep learning, data visualization, and algorithms \& complexity. Her recent work includes AI-driven reasoning systems, along with research on hybrid architectures and LLM-based pipelines.
\end{IEEEbiography}

\begin{IEEEbiography}
[{\includegraphics[width=1in,height=1.25in,clip,keepaspectratio]{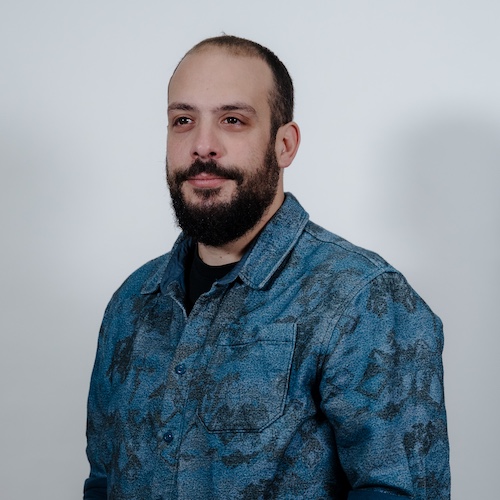}}]{Nikolaos Bakalos}  received the Diploma degree in electrical and computer engineering from the National Technical University of Athens (NTUA) in 2017. He later earned his Ph.D. from the School of Rural, Surveying and Geoinformatics Engineering at NTUA in 2022, focusing on activity recognition from visual cues.
He currently serves as a Research Engineer and Project Manager at the Institute of Communication and Computer Systems (ICCS) and NTUA. Additionally, he is an Academic Fellow at the School of Rural, Surveying and Geoinformatics Engineering, NTUA, where he teaches courses on machine learning, geoinformatics web applications, and signal processing. He has extensive experience in managing large-scale European research programs.
Dr. Bakalos has authored or co-authored over 45 scientific publications. His publication record includes 21 articles in scientific journals, 6 book chapters, and 28 papers in international conference proceedings. He also serves as a reviewer for several prestigious journals, such as the IEEE Journal of Selected Topics in Applied Earth Observations and Remote Sensing and Expert Systems with Applications
\end{IEEEbiography}

\begin{IEEEbiography}
[{\includegraphics[width=1in,height=1.25in,clip,keepaspectratio]{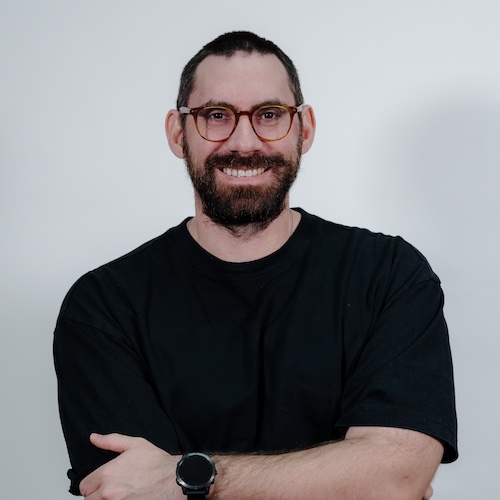}}]{Ioannis Rallis} received the diploma degree in mechanical engineering (production and management engineering) from the Technical University of Crete, Kounoupidiana, Greece, in 2014, with a thesis on forecasting methodology in finance, the postgraduate
degree in applied mathematics in modern technologies and economics from the National Technical University of Athens (NTUA), Athens, Greece, in 2016, with a focus on finance and financial engineering, and the master’s degree in investigating the the relationship between asymmetry in behavioral indicators and the stock market, and the Ph.D. degree in computer vision and machine learning for cultural heritage applications from NTUA. He has authored more than 47 publications. He has participated in more than 20 European and national research projects (H2020, Erasmus+, Interreg). His research interests include applications of machine and deep learning in an interdisciplinary field (e.g., intangible cultural heritage, remote sensing, health, finance).
\end{IEEEbiography}

\begin{IEEEbiography}
[{\includegraphics[width=1in,height=1.25in,clip,keepaspectratio]{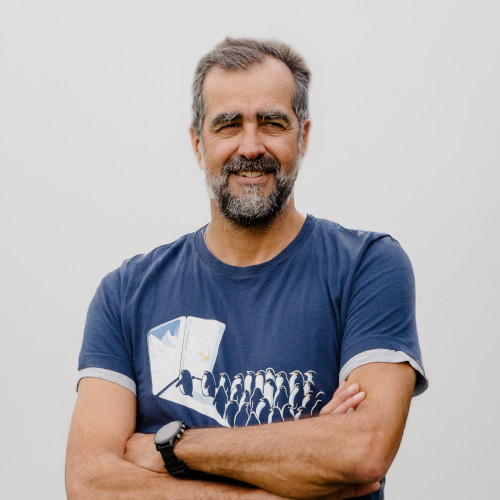}}]{Dr. Dimitris Kalogeras} is a research director with ICCS. He obtained his Ph.D. (1996) Electrical \& Computer Engineering from NTUA. His research spans several aspects of advanced network technologies and protocols. He served as a senior consultant of GRNET (the Greek NREN) and the GSN NOC. He was involved in several European RTD projects and Network Security projects. His latest research interest include  privacy-aware distributed employment of ML techniques for general cyber security and further application to medical devices. He has multiple security publications on magazines and conferences for DDOS detection and mitigation with dataplane techniques. His interests include AI/ML for decentralised security problems.
\end{IEEEbiography}
\begin{IEEEbiography}
 [{\includegraphics[width=1in,height=1.25in,clip,keepaspectratio]{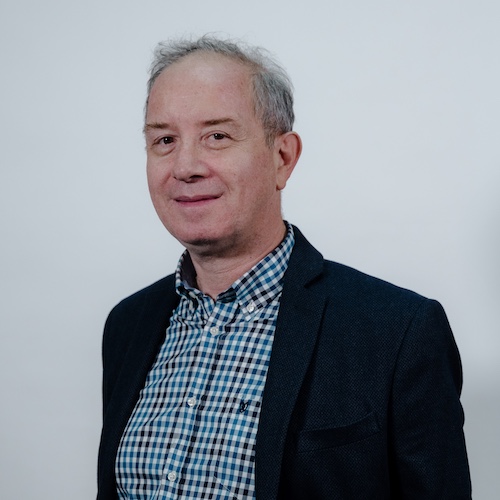}}]{Prof. Anastasios Doulamis} 
has received the Diploma degree in Electrical and Computer Engineering from the National Technical University of Athens (NTUA) with the highest honor (first ranked among all classmates) and the PhD degree in Electrical and Computer Engineering from NTUA. He is currently Professor at the National Technical University of Athens. 
Prof. Anastasios Doulamis has received several awards and prizes during his studies, including the Best Greek Student in all fields of engineering in national level, the Best Graduate Thesis Award in the area of Electrical Engineering and several prizes from the National Technical University of Athens, the National Scholarship Foundation and the Technical Chamber of Greece. He was given the NTUA Medal as Best Young Engineer. He received the best PhD thesis award from the Thomaidion Foundation. He received three best paper awards in IEEE conferences.  

He is currently the author of more than 125 journal papers, 4 books, 27 book chapters and more than 300 conference papers, the majority of which is within the IEEE archives. In NTUA, he teaches signal processing, computer vision-digital photogrammetry, programming languages and databases. 
    
He is currently involved, either as coordinator or as scientific responsible, in more than 27 Horizon Europe and Horizon 2020 European Union Projects in the broad field of remote sensing, hyper-spectral imaging for environmental and medical applications, road infrastructure inspection, computer vision methods for the preservation of cultural heritage. Examples include Horizon-PREVENT, Horizon ENVIROMED, Horizon Expidite, H2020-HEART, H2020-Gecko, H2020-Heron, H2020-Felice, H2020-Incisive.  
\end{IEEEbiography}

\begin{IEEEbiography}
 [{\includegraphics[width=1in,height=1.25in,clip,keepaspectratio]{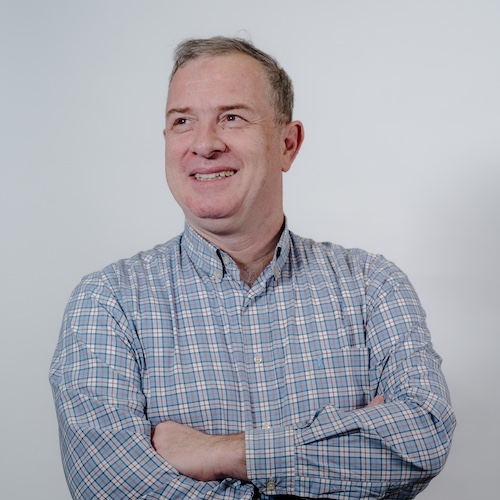}}]{Prof. Nikolaos Doulamis} has received the Diploma degree in Electrical and Computer Engineering from the National Technical University of Athens (NTUA) with the highest honor (first ranked among all classmates) and the PhD degree in Electrical and Computer Engineering from NTUA. He is currently Professor at the National Technical University of Athens. Prof. Nikolaos Doulamis has received several awards and prizes during his studies, including the Best Greek Student in all fields of engineering in national level, the Best Graduate Thesis Award in the area of Electrical Engineering and several prizes from the National Technical University of Athens, the National Scholarship Foundation and the Technical Chamber of Greece. He was given the NTUA Medal as Best Young Engineer. He received the best PhD thesis award from the Thomaidion Foundation. He received three best paper awards in IEEE conferences.  
 
He is currently the author of more than 125 journal papers, 4 books, 27 book chapters and more than 300 conference papers, the majority of which is within the IEEE archives. In NTUA, he teaches signal processing, computer vision-digital photogrammetry, programming languages and databases. 
    
He is currently involved, either as coordinator or as scientific responsible, in more than 27 Horizon Europe and Horizon 2020 European Union Projects in the broad field of remote sensing, hyper-spectral imaging for environmental and medical applications, road infrastructure inspection, computer vision methods for the preservation of cultural heritage.
\end{IEEEbiography}

\EOD

\end{document}

%% file: sections/introduction.tex
\section{Introduction}
\label{sec:introduction}

\PARstart{W}{e} live in a world full of data. Vast volumes are being generated at a pace that far outstrips our ability to organize, interpret, and extract value from it. Managing this data effectively has become one of the defining challenges of modern information systems. Structured data warehouses emerged as an early response to this challenge \cite{knezevic2025building}, \cite{hai2023datalakes}. However, their rigidity proved incongruous to the diversity and velocity of real-world data. This limitation lead to the introduction of data lakes; large-scale repositories capable of ingesting raw, unprocessed data in any format, from any source, without requiring a schema at the point of ingestion \cite{guntupalli2023data}. Data lakes may offer flexibility and scalability, but they also introduce a critical trade-off: without structure, data becomes difficult to query, integrate, and reason over. In many cases, valuable information remains unexplored or underutilized lacking a principled mechanism for connecting related data across sources. 

This challenge is particularly important in the medical domain, where huge and heterogeneous amounts of data co-exist in fragmented, schema-less repositories. Patient records, clinical notes, diagnostic reports/images, and biomedical literature are all examples of medical data spanning numerical measurements, and text and image formats thus making standardized storage and cross-source integration extremely demanding \cite{karami2017clinical}. At the same time clinical decision-making requires that information is not only stored, but semantically organized and readily accessible. It is, therefore, evident that we seek approaches capable of imposing meaningful structure on inherently schema-less medical data while preserving the flexibility that makes data lakes valuable \cite{sebaa2018medical} \cite{sukumar2015quality}. 

Knowledge Graphs (KGs) are a compelling paradigm for addressing this need. By representing knowledge as a collection of structured triplets (\textit{head entity}, \textit{relationship}, \textit{tail entity}), or simply $(h, r, t)$, KGs enable rich semantic modelling of complex interconnected information. They have been successfully applied in many real-world applications like recommendation systems, question answering \cite{sukhwal2024joint}, information retrieval. Their landscape has also been expanding in various types, such as encyclopedia knowledge \cite{auer2007dbpedia, suchanek2008yago, bollacker2008freebase, vrandevcic2014wikidata},  lexical knowledge \cite{miller1995wordnet, navigli2012babelnet, mitchell2018never, dong2014knowledge} common sense knowledge \cite{speer2017conceptnet}, \cite{hwang2021comet}, geographic knowledge \cite{auer2009linkedgeodata}, \cite{dsouza2023worldkg}, domain-specific knowledge \cite{bodenreider2004unified, li2020real, kertkeidkachorn2023finkg, li2020alimekg, dang2023gena, yang2024lmkg} and multimodal knowledge \cite{wang2020richpedia}. This breadth underscores the versatility of KGs as a general purpose knowledge representation framework.

Despite their promise, KGs have been shown to face many problems. Their construction has traditionally depended on expert-curated ontologies and manual annotation pipelines, making scalability and expert-dependency a persistent concern. They are further susceptible to knowledge incompleteness and noise, and their static nature makes it difficult to adapt to evolving domains \cite{li2025review, peng2023knowledge}. The rise of Large Language Models (LLMs) has opened new possibilities for addressing these shortcomings. Trained on massive corpora from diverse domains, LLMs encode broad world knowledge and have strong generalization capabilities across tasks. Recently, more and more synergies are being developed for unification of LLMs and KGs \cite{cai2024bringing, zhu2024llms, pan2024unifying} leveraging the complementary strength of both, thereby signaling a shift from rigid, rule-driven frameworks towards dynamic, adaptive systems.

Building on this convergence of LLMs and KGs, we present $H_2$, an end-to-end backend system that introduces a semantic data lake paradigm to medical data management. We pair the rigid schema-on-write approach of data warehouses with the schema-on-read logic of data lakes to create a hybrid solution. This results in two major outcomes; a) the formation of a predefined metadata model, stored in a document-based database to provide fundamental, rigid indexing and b) the creation of a dynamic Knowledge Graph (KG) that introduces a flexible layer with schema-on-read logic for rich knowledge extraction. Additionally, we propose a hybrid KG construction component that adopts both rule-based RDF triple creation and an integrated solution to expand the ground truth knowledge further through an automated LLM-assisted annotation system that assign operation-oriented labels to metadata objects. To maintain data integrity and avoid the creation of unstructured data silos or drifting into the "data swamp" failure mode as described in \cite{hai2023datalakes, azzabi2024datalakes}, a Human-in-the-Loop approach is introduced to provide essential verification of the LLM outputs.

The rest of this paper is organized as follows. Section \ref{sec:related_work} reviews related state-of-the-art research on semantic knowledge graph construction and enhancement using LLMs, highlighting how our approach builds upon and differentiates itself from existing work. Section \ref{sec:architecture} presents the overall architecture of the proposed semantic data lake, analyzing further the most novel parts of it, while revealing the end-to-end sequence of operations for new entry ingestion and metadata-based registration involving LLM-assisted enhancement. Section \ref{sec:methodology} outlines our approach to formulating the problem and evaluating the results using the proposed metrics. Section \ref{sec:evaluation} describes the experimental setup, presents the results and provides visualizations of some selected knowledge graph parts for qualitative assessment. Section \ref{sec:conclusion} summarizes the main aspects of our work reports the main results and provides a short commentary on the outcomes.

\subsection{Our contribution}
The contribution of this paper should be summarized in few, clear bullet points emphasizing the novelties achieved by this work. We propose:
\begin{itemize}
    \item an automated semantic triple generation mechanism which is based on the static relations identified in the raw metadata structure. This rule based mechanism creates the fundamental taxonomy of the defined semantic ontology.
    \item an LLM-assisted system that annotates metadata with operation-specific tags. In this way, we introduce an additional layer of descriptors regarding the operational readiness of data. In practice, this new information is translated to the corresponding RDF triples, which then form the extended taxonomy based on the updated semantic ontology schema. 
    \item a semantic data lake architecture that incorporates a hybrid metadata modelling schema. While flexibility is promoted through a knowledge graph layer built on the corresponding semantic ontology, a document based Database Layer maintains a fundamental metadata structure.
    \item a human-in-the-loop approach that enables the validation of LLM responses by a human while maintaining the integrity of the input data. In our work, we simulate this approach by utilizing an auxiliary LLM and evaluate its performance by introducing corresponding metrics.
    % \item We propose a \textbf{fully deployable} backend system that adopts modern, data-lake-oriented standards.
    % \item Our system supports the storage, indexing, and semantic knowledge extraction of diverse, \textbf{heterogeneous data sources}.
    % \item We provide a robust, \textbf{query-driven data discovery} interface to facilitate efficient information retrieval.
    % \item We propose a \textbf{hybrid architecture} that combines document-based storage with a \textbf{schema-on-read} approach, providing a structured metadata modeling solution alongside a \textbf{flexible layer for knowledge construction}.
    % \item We introduce a \textbf{rule-based RDF triple creation} process, further enhanced by a \textbf{Large Language Model (LLM) driven} solution. A Human-In-the-Loop (HIL) component guarantees stability and robustness.
    % \item We \textbf{evaluate several LLMs} from various categories and families to eliminate bias and identify the most effective model for our specific use case.
    % \item The proposed LLM-based mechanism is introduced as a \textbf{modular add-on}, without compromising system's functionality when it is disabled.
\end{itemize}

\noindent To facilitate reproducibility, the complete implementation code and dataset pre-processing scripts are publicly available on GitHub at \href{https://github.com/itzortzis/h2}{https://github.com/itzortzis/h2}.

%% file: sections/related_work.tex
\section{Related Work}
\label{sec:related_work}

% \subsection{Datalake architectures} \colorbox{BurntOrange}{Agapi:}  

% \subsection{LLM enhanced Knowledge graphs}

Recent research has established a taxonomy for LLM-enhanced knowledge graphs consisting of five primary categories \cite{pan2024unifying}.
These range from foundational data representation, such as improving embeddings and KG construction through knowledge extraction, to operational tasks like KG completion. Furthermore, LLMs act as a conversational layer, facilitating both the generation of natural language from structured facts (KG-to-text) and the retrieval of information through natural language question-answering systems.

In our work, we will focus only on the knowledge extraction process for the construction of a KG, which focuses on identifying and extracting entities, relations, and attributes from semi-structured and unstructured data. The knowledge extraction consists of 3 main tasks: Named Entity Recognition (NER), Relation Extraction and Linking (entity linking and coreference resolution).

\subsection{LLM-assisted Knowledge extraction: Named Entity Recognition task}

Named Entity Recognition (NER) refers to identifying and classifying entities in unstructured data. The advent of transformer architectures \cite{vaswani2017attention}, \cite{liu2019roberta}, has significantly enhanced the linguistic knowledge and contextual understanding of LLMs, making them superior in NER tasks and widely used in state-of-the-art approaches. 

GPT-NER \cite{wang2025gpt} manages to adapt LLMs to the NER task, by converting the sequence labeling task to a generation task. SF-GPT \cite{sun2025sf}, initiating from an entity extraction filter, proposes a three-step pipeline for extracting semantic triples to tackle the lack of semantic richness and interpretability in complex data. PGD-GP \cite{10460123} employs a NER model to identify and classify specific tokens in the input text with the aim of extracting meaningful relations to construct a Knowledge Graph based on predefined ontology rules. In LTNER \cite{yan2024ltner}, the authors present the Contextualized Entity Marking Gen Method in an attempt to adapt the NER task to the generative capabilities of LLMs. Accordingly, \cite{zaratiana2024gliner} and \cite{NEURIPS2024_d5aed68f} introduce validated approaches that utilize LLMs for NER tasks via well-structured prompting. Finally, studies have begun to explore multi-modal NER (MNER) within the context of KGs. The main objective in such cases is to map visual content, i.e. images, to the respective entities in the textual sources \cite{zhang2024mkeah}, \cite{li2020gaia}. 

The common ground of the aforementioned works relies primarily on utilizing LLMs for token-based identification and classification, with some frameworks providing additional functionality by extracting semantic triples based on ontological rules. We were motivated by these publications, which practically demonstrate how generative LLMs can be leveraged to label identified objects. Furthermore, in our case, we formulate our task as a multi-label classification problem and, rather than following a token-based approach, we employ document-based categorization. Following this approach, our novelty relies on the creation of an automatic annotation system which assigns operation-related labels to identified entities based on the corresponding metadata document.  

% Eventually, our ultimate goal is to extend the rule-designed taxonomy of our ontology by creating new RDF triples based on the LLM suggestions.

\begin{figure*}[!h]
    \centering
    \includegraphics[width=1\linewidth]{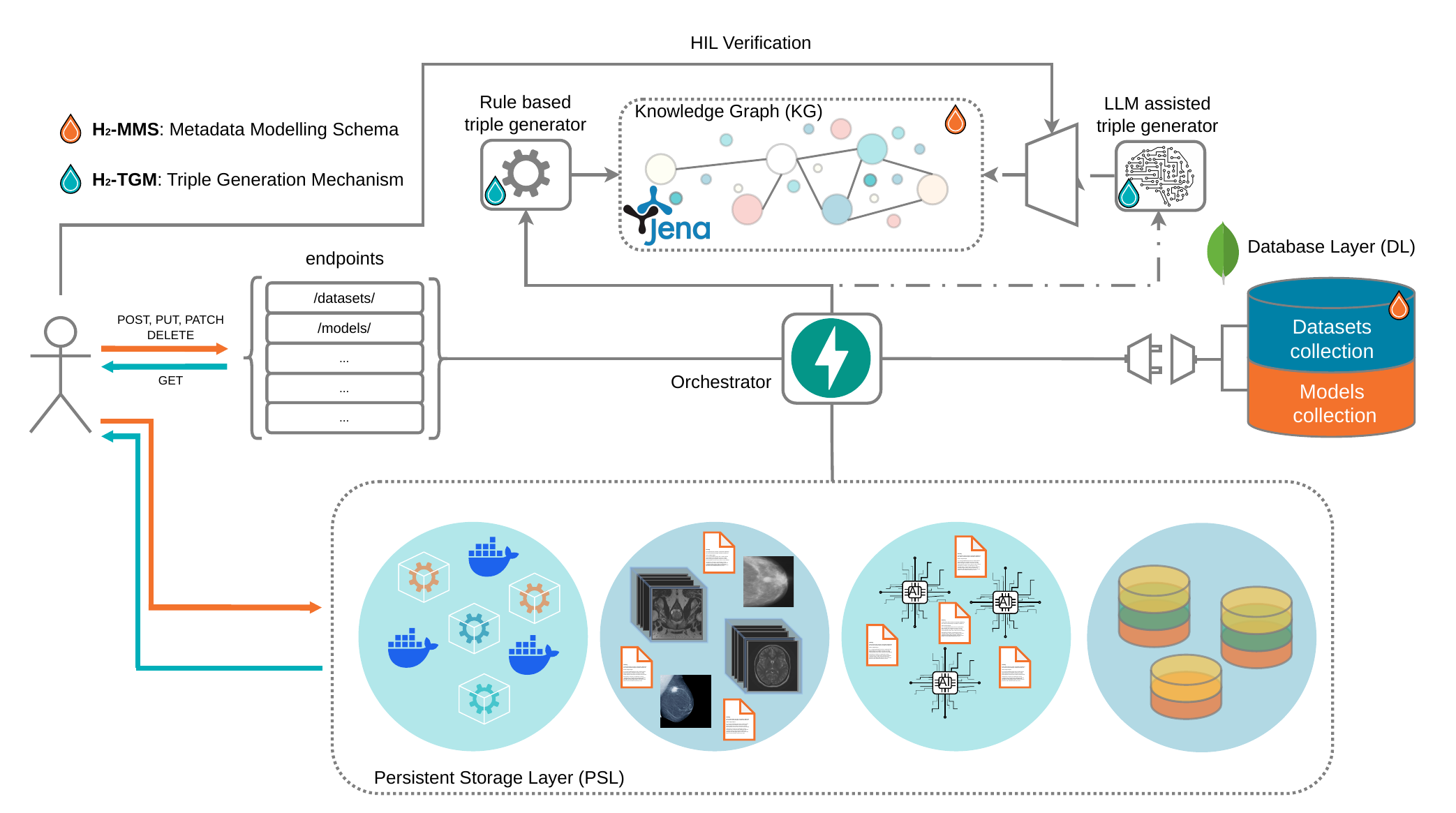}
    \caption{The proposed $H_2$ architecture. The two main novelties are highlighted: a) the Document-based and Schema-on-Read approach introduced by the $H_2$-$MMS$ module and b) the Rule-based semantic triple generator and LLM-driven metadata annotation mechanism implemented by the $H_2$-$TGM$ module. The output of the latter is verified by the proposed Human-In-the-Loop (HIL) solution. }
    \label{fig:datalake_arch}
\end{figure*}

\subsection{LLM-assisted Knowledge extraction: Relation Extraction task}
Relation extraction refers to recognizing the relationships between the identified entities within a text.

BertNet \cite{hao2023bertnet} is a novel automatic framework that allows the automatic KG construction using a pretrained LLM (e.g, BERT, ROBERTA). It requires a minimal input of a relation definition: a prompt accompanied by a few entity pairs examples. A search-and-rescore mechanism is also applied to improve both efficiency and accuracy. ATLOP \cite{zhou2021document} focuses on solving multi-label and multi-entity problems, by using an enhanced BERT-baseline and incorporating two novel techniques: adaptive thresholding and localized context pooling. The former utilizes a learnable entities-dependent threshold, instead of the global threshold. The latter leverages pre-trained attention heads for the localization of relevant context for entity pairs. DREEAM \cite{ma2023dreeam} proposes a memory-efficent approach, that incorporates ATLOP as its backbone and further improves it, by adopting evidence information as the supervisory signal. In \cite{li2026llm} they introduce a classifier-llm approach for document-level relation extraction (DocRE), which constists of two steps. Firstly, entity pair candidates are identified through a classifier, which are then fed to an LLM to determine the final relations. Similarly, \cite{zhong2025leveraging} also contributes in DocRE, by proposing a two-stage framework, which first extracts and classifies relation triplets candidates using a BERT-based model, and then an LLM is utilized to perform the final relation classification. 

The works presented in the publications of this subsection address the identification of relations among entities within an entire document. Even though their backend mechanisms may boil down to token-based approaches, as discussed in the previous subsection, their high-level methodology operates at the document level. These approaches leverage LLMs to identify potential relationships in the input document and form the corresponding semantic triples. In our case, we adopt this document-level approach by prompting LLMs to assign the given context to predefined labels. In contrast to the aforementioned approaches, our solution promotes a more structured construction of semantic triples, by constraining the available label options to a specific target list and setting predefined operation-oriented relation predicate.
% To ensure knowledge integrity, rather than allowing the LLM to freely generate semantic triples, we define the target relationships and ask the model to classify the context accordingly.
%if we want to mention survey:
% \cite{diaz2025survey} and \cite{zhao2024comprehensive}

\subsection{LLM-assisted Knowledge extraction: coreference resolution and entity linking}
Entity Linking refers to both coreference resolution and entity linking. The former involves linking the various mentions of a specific entity or event in a text. The latter, involves creating a link between the identified entities and the KG. 

% coreference resolution

% In \cite{joshi2019bert}, the authors improved \cite{lee2018higher}'s work for coreference resolution, by replacing the entire LSTM-based encoder with the BERT transformer. Later on, this work was further extended by introducing SpanBERT \cite{joshi2020spanbert} and using it instead of the BERT.
% The computation time and memory load of coreference resolution were then improved by the work of \cite{kirstain2021coreference}, who proposed a start-to-end (s2e) coreference model, that is not dependent on span representations. This work was then used as a baseline for the development of fastcoref \cite{otmazgin2022f}, a python package for coreference resolution and was further extended by the introduction of LINGMESS \cite{otmazgin2023lingmess}. Linguistically Informed Multi Expert Scorers (LINGMESS) is a coreference model that uses a dedicated trainable scoring function for different mention-pairs categories.

%Maverick: Efficient and accurate coreference resolution defying recent trends
%\cite{martinelli2024maverick}

%In \cite{chen2025improving} the authors enhance cross-document event coreference resolution, by %introducing a new task called Event-oriented cross-document coherence enhancement (ECD-CoE). This %method extracts coherent sentence sets for two cross-document event mentions and transforms them %into tree structures with rhetorical relations, to finally extract the coreference relation.

% entity linking

The authors of \cite{broscheit2019investigating} exploit a highly simplified approach of entity linking, in order to investigate how much entity knowledge is stored within BERT's pretrained representations. In the work of \cite{ayoola2022refined}, the scalable end-to-end entity linking model ReFinED is introduced, which leverages the RoBERTa architecture to achieve linking through entity types and entity descriptions. The authors of \cite{de2020autoregressive} introduced GENRE, which leverages a sequence-to-sequence architecture to generate entity names autoregressively. Later, mGENRE \cite{de2022multilingual} extended GENRE's work to its multilingual version.  Finally, UniMEL \cite{liu2024unimel} also focuses on multimodal entity linking by introducing a three-step approach. Firstly, an LLM is utilized for the individual augmentation of mentions and entities representation. Then, after filtering potential matches with an embedding-based technique, the LLM acts as a final selector to identify the entity that most accurately corresponds to the mention. 

The cited works in this domain apply not only to sentences but also to entire documents, with some providing multimodal analysis. Their common ground is the Entity Linking task, which includes identifying, classifying, and correlating entities to construct a semantic taxonomy. We propose a document-to-entity linking approach by providing candidate entities from which the LLM selects the most appropriate matches based on the given relation and input entry. Our approach enhances data linking through a Human-In-the-Loop (HIL) validation process, maintaining data integrity while providing a robust framework to expand the core taxonomy.

% To maintain data integrity, our solution is intentionally constrained, incorporating human-in-the-loop (HIL) validation for further verification of the LLM's outputs.

%% file: sections/architecture.tex
\section{System Architecture}
\label{sec:architecture}

The architecture of the proposed Knowledge Base system is presented in Figure \ref{fig:datalake_arch}. As shown, several standalone components compose the final structure of the system. The Persistent Storage Layer (PSL) is designed to store the incoming files in their raw form, without considering their internal characteristics like data type, size or format. In this sense, there is no specific structure dictated at the stage of data ingestion. The Database Layer (DL), through its document-oriented nature, indexes details about the stored data and provides a flexible way to store customized metadata depending on the needs of each entry. To maintain a fundamental consistency towards the design of a robust metadata schema, some data details are considered as required inputs and expected to be provided by the owner during the registration process. Eventually, rule-based triple generation mechanism construct the fundamental taxonomy of the KG, which is then enhanced by the interpretation of the LLM assisted triple generation module. The API based orchestrator provides the interconnection among the rest of the modules and serves as the primary interface for the system's client.

\subsubsection{$\mathbf{H_2}$\textbf{-}$\mathbf{MMS}$: Metadata Modelling Schema}
Our proposed Metadata Modelling Schema ($H_2$-$MMS$) provides a hybrid solution to index ingested data and extract relevant, actionable knowledge flexibly, without compromising the rigid organization required to prevent a data swamp. To this end, we employ custom models to establish a schema foundation and store incoming metadata systematically within the designated database collections. This flexibility is enabled by the ability a) to add, alter, or remove defined models; b) to instantiate models with varying attributes via optional field definitions; and c) to adjust the rules dictating how the main RDF triples are generated from the raw metadata. As depicted in Figure \ref{fig:datalake_arch}, the $H_2$-$MMS$ module includes both the Knowledge Graph (KG) and the MongoDB document-based collections, each marked with the orange drop-like icon.

\subsection{Proposed dual LLM component for RDF triple generation}

% Using the constructed datasets, all the matching to the target list tasks of the entries were extracted and formed as ground truth labels. Each dataset entry is filtered in order to subtract the ground truth tags while the rest of the information is passed to LLM prompt as shown in Figure \ref{fig:generator_prompt}.
As shown in Figure \ref{fig:generator_prompt}, the LLM is defined to be an AI expert, specialized in dataset categorization and is instructed propose task-related tags based on the given title, description and tags. Some additional rules are also defined aiming to form the model's output as well as possible.

\subsubsection{The triple generator LLM}
The generator LLM expects to receive the required metadata of the new entry. In our case those are the title, the description and the tags of the entry. As shown in Figure \ref{fig:generator_prompt}, the model is directed to identify 1-3 ML tasks that appear to best fit the current entry. To limit the model "imagination" and prepare a more stratified output, a target list is included in the prompt and contains all the available options the model is expected to select from. Finally, some procedural rules along with some example outputs are provided as well in order to properly form the output of the model. During the experimentation phase, tags given to the generator were filtered so as to be absent from the target list. In doing so, we established a ground truth to evaluate the model's subsequent performance.

\subsubsection{Triple evaluation through simulated Human-in-the-Loop  LLM (HIL-LLM)}
The secondary LLM is introduced to simulate the HIL as it is described in section \ref{sec:workflow}. As appears in Figure \ref{fig:human_prompt}, the same role has been assigned according to which the HIL-LLM is an AI expert specializing in dataset categorization tasks. In this case, not only the full tag list is given as input (without filtering out ground truth labels), but also the generator's output is provided. Thus, this LLM is directed to compare the original tags with the generator's proposals and output a list with the matches.

\input{prompts/generator}

\input{prompts/human}

\subsubsection{$\mathbf{H_2}$\textbf{-}$\mathbf{TGM}$: Triple Generation Mechanism}
The Triple Generation Mechanism ($H_2$-$TGM$) refers to the module responsible for creating new RDF triples to enhance the KG. This module introduces the second hybridity of the proposed system, not only by leveraging a rule-based mechanism to extract meaningful RDF triples from the raw metadata, but also by adopting an LLM-assisted approach to enrich the existing knowledge by applying automated operation-oriented annotation to the new metadata entries. As shown in Figure \ref{fig:datalake_arch}, this module consists of the \textit{Rule-based triple generator} and the \textit{LLM-assisted triple generator} components, each marked with the turquoise drop-like icon.

% \textcolor{red}{H2-MMS and H2-TGM should be added as subtitles.}
% \textcolor{red}{The following paragraph should be replaced}
% Aiming to design a flexible datalake system that supports heterogeneous data, it is not difficult to end up implementing a stash-like storage. To avoid such a situation...Before diving into the system architecture, it is important to explain the registration process that follows any new data entry, to prevent the creation of a stash-like datalake. Upon completion of the file upload, the user should provide textual details about the new entry, forming metadata that can be used to index the corresponding data. This raw metadata is then stored in the database and constitutes the basis for the expansion of the Knowledge Graph. This data indexing process, which spans from the collection of details to the generation of new RDF triples, is referred to as registration. In this work, we attempt to enhance the final registration step by employing an LLM to extract useful information from the provided metadata, thereby further enriching the Knowledge Graph with a Human-In-the-Loop approach.

% \begin{figure*}[!t]
%     \centering
%     \includegraphics[width=0.9\linewidth]{figs/architecture.pdf}
%     \caption{The proposed $H_2$ architecture. The two main novelties are highlighted: a) the Document-based and Schema-on-Read approach introduced by the $H_2$-$MMS$ module and b) the Rule-based and LLM-driven mechanism implemented by the $H_2$-$TGM$ module.}
%     \label{fig:datalake_arch}
% \end{figure*}

\subsection{Proposed End-to-end workflow for new entry registration } \label{sec:workflow}

% To enhance the discoverability of knowledge triples for the datasets, we developed a Human-in-the-Loop (HITL) enrichment pipeline that bridges LLM predictions with expert human validation. At the beginning, an LLM is utilized to perform zero-shot classification of datasets into specific machine learning tasks (e.g., segmentation, regression) based on their metadata. Then, a human operator interacts with the system to either accept these suggestions or manually contribute additional domain-specific tags. Once the final set of tags is established, the rdflib library formalizes the entries into RDF triples.
% These are then pushed to an Apache Jena Fuseki triplestore via the Graph Store Protocol, and are integrated into a broader knowledge graph.

\begin{figure*}[!h]
    \centering
    \includegraphics[width=1\linewidth]{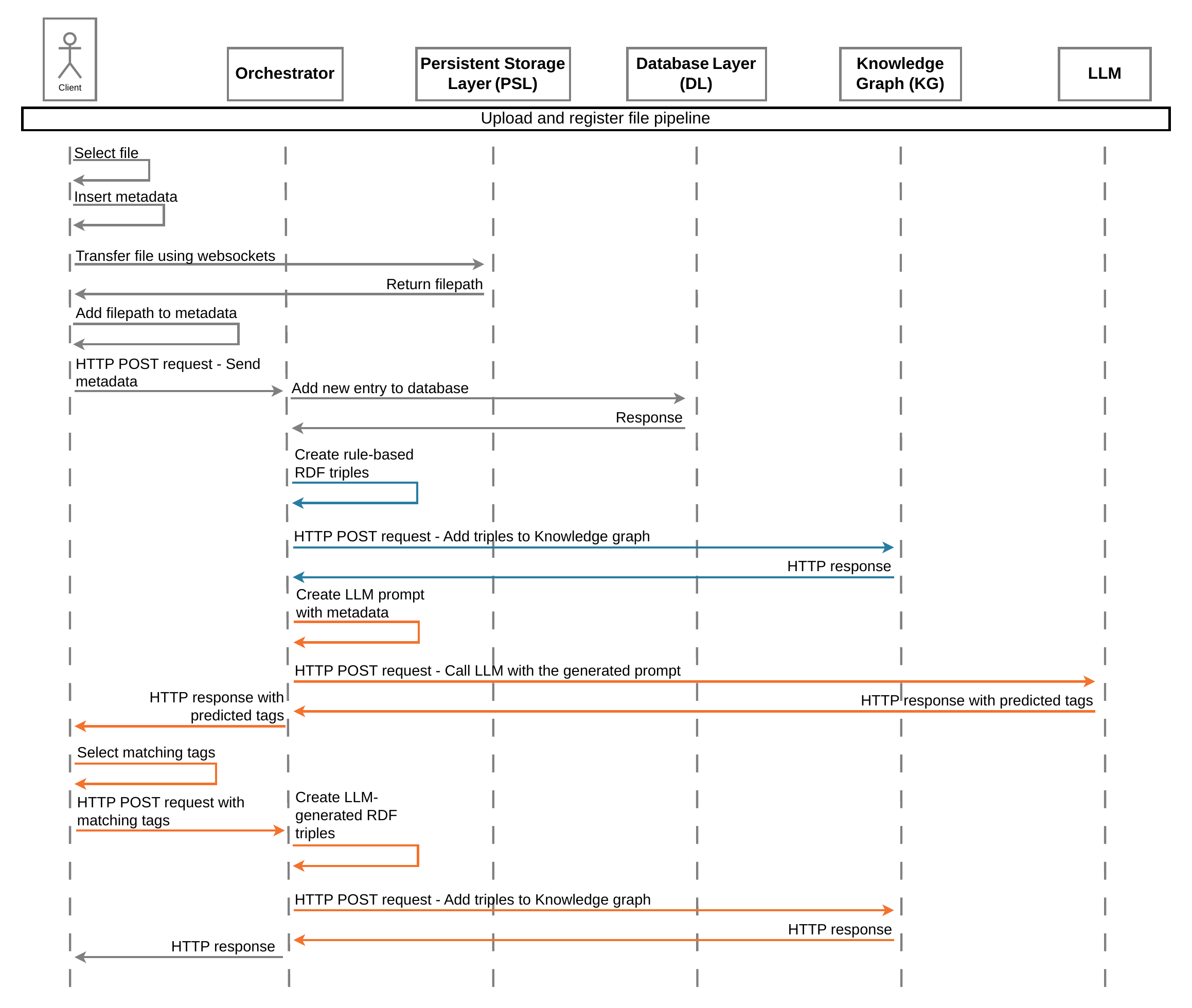}
    \caption{End-to-end sequence of operations for new entry ingestion and registration involving LLM-enabled metadata annotation.}
    \label{fig:sequence}
\end{figure*}

The sequence diagram in Figure \ref{fig:sequence} provides a tangible description of the actions taking place to add a new entry to the proposed data lake system, from the inputs of the clients, the intercommunication of the internal components, the LLM enhancements, to the creation of the final KG entry. At first the client software establish an appropriate communication with the server program of the PSL responsible for accepting and organizing the incoming data. Once the ingestion procedure is completed, the server program returns the corresponding response that includes the exact location of the new entry in the filesystem. Upon the collection of the essential information, An HTTP POST request is employed by the client software to send the metadata to the API of the system, which works as an orchestrator for the further actions. It creates an appropriate request to the DL aiming to store the raw metadata there. Once a successful response has arrived, the orchestrator generates the basic RDF triples from the given metadata using predefined rules and creates the corresponding HTTP POST request to the KG of the system in order to push these new entries. At this points the API utilizes the raw metadata to format a prompt template and attempts a call to the LLM API. The latter gets triggered, processes the request and shapes the corresponding response with the answer included. The API checks the answer and if is of a proper form it returns it as response to the client. Upon the selection of the matching tags, the client creates a new HTTP POST requests with the selected tags included. The API again creates the new RDF triples and attempts a new request to add them to the Knowledge graph.

\begin{figure*}[!h]
    \centering
    \includegraphics[width=1\linewidth]{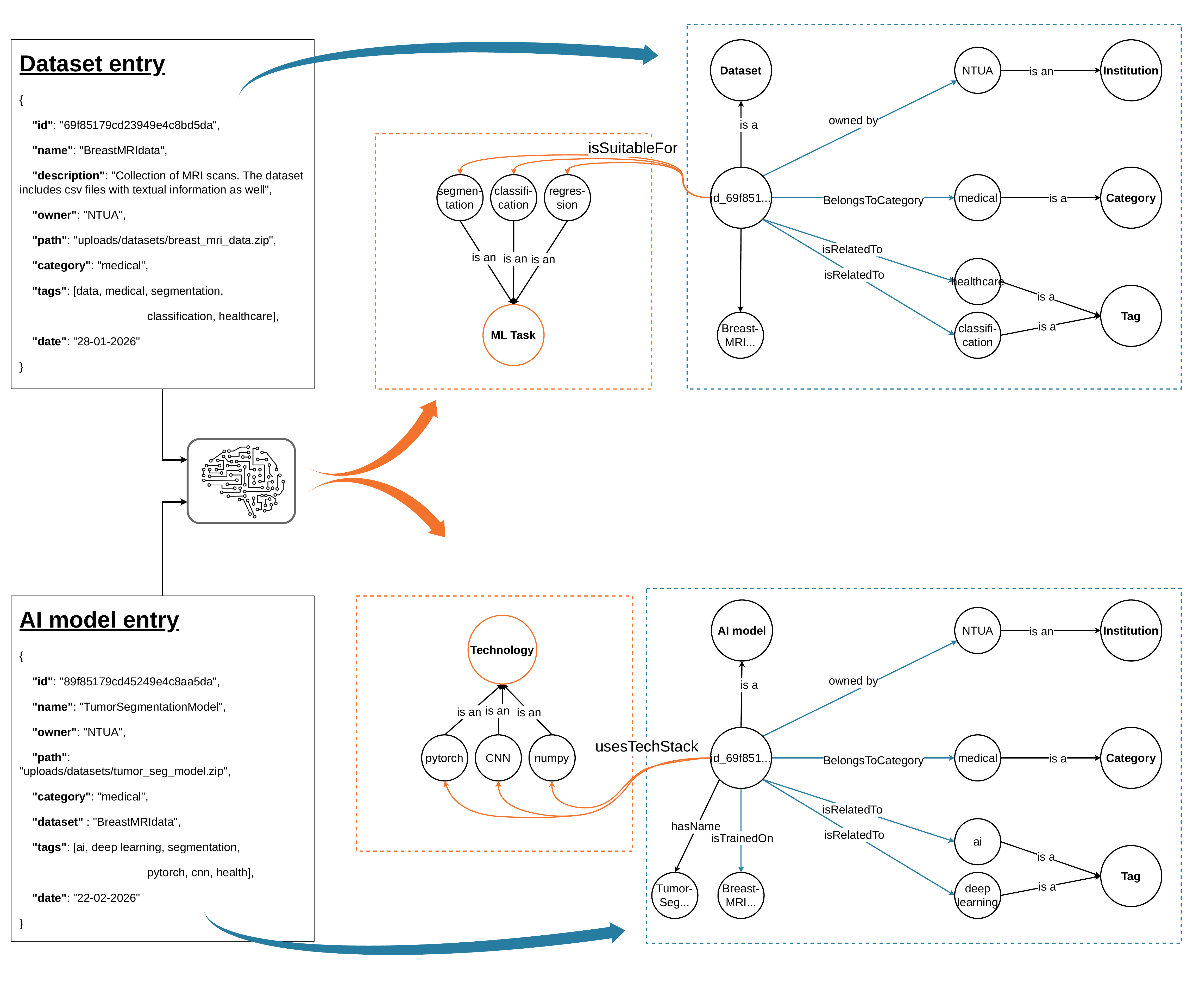}
    \caption{Two simplified paradigms illustrating the creation of the base and extended taxonomy. Blue arrows represent the automatic translation of raw metadata to the base taxonomy, while orange arrows show its extension using LLM-generated, AI operation-oriented tags.}
    \label{fig:usecases}
\end{figure*}

In Figure \ref{fig:usecases}, two simplified paradigms are presented where we showcase the automatic generation of a) the base taxonomy according to the predefined semantic ontology and b) the extended taxonomy provided by the LLM based metadata annotation system. In the first case, the new entry corresponds to a dataset metadata object that includes the following fields: $id, name, description, owner, path, category, tags, and date$. Of these, $id, name, owner, and category$ are part of the semantic ontology definition, so they are considered required fields. According to our semantic model, $id$ and $owner$ are connected via the $ownedBy$ relation, where the former is expected to be a $Dataset$ and the latter an $Institution$, forming, eventually, the semantic triple $Dataset \mapsto ownedBy \mapsto Institution$. Accordingly, $belongsToCategory$ connects this dataset with the category "medical", forming the semantic triple $Dataset \mapsto belongsToCategory \mapsto Category$. Even though $tags$ field in not required, when it exists, our solution creates relations denoted by $isRelatedTo$ predicate for each tag in the list. Like other fields as well, $name$ constitutes an attribute for the $Dataset$ entity. Due to its high importance, $description$ field is also set as required since it consists of the owner's raw description which may include valuable information about the related AI operations. After the creation of the base taxonomy, the whole metadata object is passed to the LLM which is then tasked to assign AI operation related tags (i.e. classification, regression, segmentation, etc) from a given list that match the given context. After the verification of the generated tags, our proposed system creates the corresponding semantic triples using the relationship $isSuitableFor$ using the dataset identifier as subject and each task as the corresponding object. The second use case follows the same sequence for an AI model entry. The main difference lies in the semantic relationships the LLM identifies based on labels representing the training and inference technology stack. After verifying the selected technology-related tags, our proposed system connects the model's identifier to each tag via the $usesTechStack$ relationship.

%% file: prompts/generator.tex
\begin{figure}[ht]
\centering
\hrule \vspace{2pt} \hrule
\vspace{5pt}
\begin{minipage}{0.95\columnwidth}
\footnotesize \ttfamily
\noindent You are an \textbf{AI expert} specializing in dataset categorization. \\
\\
\textbf{DATASET INFO:} \\
Title: "\{title\}" \\
Description: "\{subtitle\}" \\
Existing Tags: \{tags\} \\
\\
\textbf{TASK:} \\
Identify 1-3 Machine Learning tasks from the \textbf{TARGET LIST} that best fit this dataset. \\
\\
\textbf{TARGET LIST:} \\
\{target\_tasks\} \\
\\
\textbf{OUTPUT RULES:} \\
1. Output MUST be a valid Python list of strings. \\
2. Use ONLY tags from the TARGET LIST. \\
3. Maximum 3 tags. \\
4. NO keys, NO explanations, NO markdown blocks. \\
\\
\textbf{EXAMPLE OUTPUT:} \\
{[}'binary\_classification', 'image\_segmentation'{]}
\end{minipage}
\vspace{5pt}
\hrule \vspace{2pt} \hrule
\caption{Structure of the generator-LLM prompt, highlighting the input, task definitions, rules, and output format.}
\label{fig:generator_prompt}
\end{figure}

%% file: prompts/human.tex
\begin{figure}[ht]
\centering
\hrule \vspace{2pt} \hrule
\vspace{5pt}
\begin{minipage}{0.95\columnwidth}
\footnotesize \ttfamily
\noindent You are an \textbf{AI expert} specializing in dataset categorization. \\
\\
\textbf{DATASET INFO:} \\
Title: "\{title\}" \\
Description: "\{subtitle\}" \\
Existing Tags: \{tags\} \\
\\
\textbf{PROPOSED TASKS:} \\
\{proposed\_tags\} \\
\\
\textbf{TASK:} \\
Based on the title, the description and the existing tags, output the proposed tags that match. \\
\\

\textbf{OUTPUT RULES:} \\
1. Output MUST be a valid Python list of strings. \\
2. Use ONLY tags from the PROPOSED TASKS. \\
3. NO keys (like "new\_tags:"), NO explanations, NO markdown blocks. \\
4. If no tags are appropriate, output an empty list []. \\
\\
\textbf{EXAMPLE OUTPUT:} \\
{[}'binary\_classification', 'image\_segmentation'{]}
\end{minipage}
\vspace{5pt}
\hrule \vspace{2pt} \hrule
\caption{Structure of the simulated Human-in-the-Loop
LLM (HIL-LLM) prompt, highlighting the input, task definitions, rules, and output format.}
\label{fig:human_prompt}
\end{figure}

%% file: sections/methodology.tex
\section{Methodology}
\label{sec:methodology}

In this section, we present the end-to-end procedure, from the dataset collection process, the problem formulation, the primary LLM setup, the HIL module implementation and the evaluation metrics definition.

% \subsection{Datasets}

\subsection{Problem Formulation}
To formulate our method, let $D$ denote our dataset which consists of $(X_i, T_i)$ pairs, where $X_i$ represents the metadata describing the $i_{th}$ entry and $T_i$ the corresponding original tags (e.g. "dataset", "medical", "classification", "breast", "MRI", etc) assigned to it. Assuming N entries, the set $D$ is defined by Equation \ref{eq:dataset}

% To mathematically formulate our method, we treat the task as a multi-label classification problem, where a model selects multiple labels from a predefined target list for each entry. Let us denote as $X_i$ the metadata describing the $i_{th}$ entry and $T_i$ the corresponding original tags assigned to it. As shown in Eq. \ref{eq:dataset}, $(X_i, T_i)$ form our dataset $D$.
\begin{equation} \label{eq:dataset}
    D = \{(X_i, T_i)\}_{i=1}^N
\end{equation}

\noindent Let $\mathcal{L}$ be the target set containing all available tags (e.g. "regression", "classification", "segmentation", etc):

\begin{equation}
    \mathcal{L} = \{\lambda_1, \lambda_2, \dots, \lambda_k\}
\end{equation}

\noindent where $\lambda_i$ denotes the $i$-th tag in the set for $i \in \{1, \dots, k\}$. For a specific entry $j$, we define two subsets of $\mathcal{L}$:

\begin{itemize}
    \item The \textbf{ground truth} set $Y_j \subseteq \mathcal{L}$. This set is derived by taking the intersection of the entry's original tags $T_j$ and the target set: $Y_j = T_j \cap \mathcal{L}$.
    
    \item The \textbf{predicted} set $\hat{Y}_j \subseteq \mathcal{L}$, which is generated by the LLM subject to the cardinality constraint $|\hat{Y}_j| \leq 6$.
\end{itemize}

\noindent Both ground truth and predicted sets are mapped to binary label vectors $\mathbf{y}_j, \mathbf{\hat{y}}_j \in \{0, 1\}^k$. The elements of these vectors are defined using the indicator function $\mathds{1}$:

\begin{equation}
    y_{ji} = \mathds{1}(\lambda_i \in Y_j)
\end{equation}

\begin{equation}
    \hat{y}_{ji} = \mathds{1}(\lambda_i \in \hat{Y}_j)
\end{equation}

\noindent where $y_{ji}=1$ if the $i$-th tag is present in the ground truth for entry $j$, and $0$ otherwise.

Eventually, we treat the constrained generation task as a multi-label classification problem prompting the LLM to assign all the appropriate labels in $L$ to a given input $X_i$. Then, we compare the predictions with the corresponding ground truth labels to calculate the proposed metrics.

\subsection{Evaluation Metrics}
To evaluate the performance of the LLM across the entire dataset of $N$ entries, we employ micro-averaged metrics. This approach aggregates the counts of True Positives ($TP$), False Positives ($FP$), and False Negatives ($FN$) globally:

\begin{equation}
    TP_{total} = \sum_{j=1}^{N} |Y_j \cap \hat{Y}_j|
\end{equation}

\begin{equation} \label{eq:recall}
    Recall_{micro} = \frac{TP_{total}}{\sum_{j=1}^{N} |Y_j|}
\end{equation}

\begin{equation} \label{eq:precision}
    Precision_{micro} = \frac{TP_{total}}{\sum_{j=1}^{N} |\hat{Y}_j|}
\end{equation}

\noindent This metric choice ensures that the evaluation reflects the overall density of correctly identified tags across the complete set of metadata entries.

Following the proposed HIL approach, we introduce additional 'soft' metrics calculated based on the HIL-LLM decisions. This model is tasked with identifying which tags proposed by the generator LLM align with the metadata of the corresponding entry. It outputs a list of accepted tags, which is then concatenated with the ground-truth tags to construct an extended ground-truth list. Formally, let $\hat{Y}_j^a \subseteq \hat{Y}_j$ be the set of proposed tags approved by the HIL-LLM and let $Y_{ext}$ represent the extended ground-truth list. The 'soft' metrics are then calculated as shown in Equations \ref{eq:recall}, \ref{eq:precision} by using the new $TP^s_{total}$ in Equations \ref{eq:gtextended} - \ref{eq:tpsoft}.

\begin{equation} \label{eq:gtextended}
    Y_j^{ext} = \hat{Y}_j^a \cup Y_j
\end{equation}

\begin{equation} \label{eq:tpsoft}
    TP_{total}^s = \sum_{j=1}^{N} |Y_j^{ext} \cap \hat{Y}_j|
\end{equation}

% \begin{equation} \label{eq:recallsoft}
%     Recall_{micro}^s = \frac{TP_{total}^s}{\sum_{j=1}^{N} |Y_j|}
% \end{equation}

% \begin{equation} \label{eq:precisionsoft}
%     Precision_{micro}^s = \frac{TP_{total}^s}{\sum_{j=1}^{N} |\hat{Y}_j|}
% \end{equation}

%% file: sections/results.tex
\section{Experimental Evaluation}
\label{sec:evaluation}
\subsection{Experimental Setup}
Suitable datasets for testing our framework were identified by employing data science hubs like Kaggle to create metadata catalogues. Among other details, the metadata include information about the name, the description, the author and the given tags of each entry, which represents a publicly avalable dataset or a model. To extract health-related information, keywords like "medical" or "cancer" were adopted as search terms of the corresponding API calls. At a second stage, a list of target keywords (e.g. ML\_tasks: ["image segmentation", "classification", "regression"]) was used in order to filter out the entries in which none of these keywords were included. Following this approach, multiple datasets were generated and served as a proof-of-concept basis for testing the proposed framework.

To create a rigid basis for the experimental setup, we created four different metadata datasets based on two parameters: the search term and the data type. Let $D_1$ denote the dataset obtained using the search term 'medical' and the type 'datasets'; $D_2$, the dataset generated using 'cancer' and 'datasets'; and $D_3$, the dataset obtained using 'health' and 'models'.

In parallel, to achieve variety regarding the employed LLMs, we introduce seven different models to evaluate their performance on the aforementioned datasets. The models are categorized into three distinct tiers based on their parameter counts: the low tier, comprising models with fewer than 10 billion parameters; the medium tier, including models with 10 to 50 billion parameters; and the high tier, consisting of models with more than 50 billion parameters. To avoid any biased results occur by training philosophy, we selected models from three different families; a) Gemma (Google) - Represents the latest in Transformer-based efficiency, particularly with the Gemma 3 release, which emphasizes high-reasoning capabilities in a compact footprint \cite{team2024gemma}, b) Mistral/Ministral (Mistral AI)- Known for optimized attention mechanisms (like Sliding Window Attention) and strong performance in structured data tasks, c) Llama (Meta) - Chosen as the industry-standard open-weights baseline, providing a benchmark against which newer or more specialized architectures are measured \cite{grattafiori2024llama}.

\subsection{Performance evaluation}
In our comparisons, the key metric is $Recall$, as it represents the percentage of ground-truth tags correctly identified by the LLM. $Precision$ is also an important metric, indicating the accuracy of the model's output relative to the ground truth. A high $recall$ combined with low $precision$ implies that the model has successfully identified the ground-truth tags but has also proposed additional ones. In fact, this is the ideal outcome for our purposes since ground-truth tags provided by data owners are often limited and LLM-assisted enhancement becomes essential for a more comprehensive metadata profile. However, this enhancement must be managed to filter out failures, such as model hallucinations. To ensure a robust system, we incorporate a Human-in-the-Loop (HIL) approach into the pipeline. During experimentation, we utilized an LLM to simulate this HIL component, allowing us to calculate 'soft' metrics. In this context, both soft recall and soft precision are expected to be high, as the expanded ground-truth list is now refined through simulated human intervention.

\begin{table}[h]
\centering
\caption{Summary of metadata collections}
\label{table:datasets-description}
\begin{tabular}{@{}ccccc@{}}
\toprule
\textbf{Dataset} & \textbf{Search Term} & \textbf{Data Type} & \textbf{\# Entries} & \textbf{Task}   \\ \midrule
$D_1$            & 'medical'            & 'datasets'         & 500                & 'isSuitableFor' \\
$D_2$            & 'cancer'             & 'datasets'         & 500                & 'isSuitableFor' \\
$D_3$            & 'health'             & 'models'           & 140                & 'usesTechStack' \\ \bottomrule
\end{tabular}
\end{table}
 
In the low-tier category, Gemma3:4B consistently outperforms the other models, except in the $D_3$ case where Llama3.1:8B appears superior. Similarly, Gemma3:27B surpasses all other models across all datasets in the medium tier. Finally, Llama3:70B is the sole representative of the high-tier category and is therefore the top performer. The interesting part to be mentioned is that Gemma3:4B presents  the highest performance across tiers as well, with the exeption of $D_3$ case. This implies that for metadata extraction and classification tasks (like "$isSuitableFor$ MLtask:"), the number of model parameters is not the sole determinant of success. 

Tables \ref{table: D1} - \ref{table: D3} correspond to the evaluation metrics calculated on the datasets $D_1, D_2, D_3$ respectively. In each table, the tiers are defined by horizontal lines grouping together the sibling models. So, ministral3:3B, gemma3:4B, mistral:7B and llama3.1:8B belong to the low tier, ministral3:14B and gemma3:27B constitute the medium tier while llama3:70B itself represents the high tier.

During experimentation, for datasets $D_1$ and $D_2$, the LLMs were tasked with identifying potential ML tasks for which each entry (dataset) could be suitable. For dataset, $D_3$, the LLMs were prompted to specify the technologies adopted to construct each entry (model). 
Table \ref{table:datasets-description} summarizes the characteristics of each dataset and how it was used in the proposed pipeline based on the assigned task of the LLM.

Despite its tiny amount of parameters compared to the Llama3.1:70B, Gemma 3 4B is a much newer model, released in early 2025, and incorporates more modern architectural and training approaches. One of its key innovations is a hybrid local–global attention mechanism, which improves efficiency and enables the model to handle longer context lengths with lower memory overhead, resulting in faster inference. In this way, Gemma3:4B maintains semantic coherence without the computational overhead of high-tier models. The Llama 3.1-8B's leading performance in the $D_3$ case indicates a shift in task complexity. This suggests the model is better equipped to identify technology stacks, which requires deeper tool-reasoning capabilities compared to the ML tasks in $D_1$ and $D_2$ that are often resolved through simple vocabulary processing. As noted in \cite{grattafiori2024llama}, the Llama 3 training mixture includes a significant emphasis on reasoning and tool use. Therefore, such an outcome is consistent with its architectural design.

In an attempt to further investigate the experimental results and gain a clearer understanding of the models' performance, we present the radar plots in Figures \ref{fig:d1_results}, \ref{fig:d2_results}, and \ref{fig:d3_results}, which correspond to the results of the experiments conducted on $D_1$, $D_2$, and $D_3$, respectively. To maintain consistency and promote readability, all figures are designed in a unified way: a) each radar plot displays the metrics recall, speed, soft-precision, soft-recall, and precision in counterclockwise order; b) the large radar plot depicts the performance of all utilized models; c) each small radar plot refers to a single model; d) low-tier models are represented by blue, medium-tier models by orange, and high-tier models by green; and e) the variance of the results is denoted by a semi-transparent area. It is not difficult to observe that the shapes in Figures \ref{fig:d1_results} and \ref{fig:d2_results} are similar and the corresponding values are close, fact that is not so obvious when analyzing the raw numbers. This occurs due to the similarity of $D_1$ and $D_2$ structure and content. On the other hand, the shapes in Figure \ref{fig:d3_results} are differentiable since the dataset refers to AI models and the models are tasked with a different prompt. This occurs due to the structural and content similarities between $D_1$ and $D_2$ while, on the contrary, the shapes in Figure \ref{fig:d3_results} are distinct because that dataset focuses on AI models tasked with a different prompt. 

% \begin{table*}[!h]
% \centering
% \resizebox{\textwidth}{!}{
% \begin{tabular}{lccccc}
% \hline
% Model & Recall & Precision & Soft Recall & Soft Precision & Mean LLM time \\
% \hline
% ministral-3-3b & $0.579 \pm 0.012$ & $0.214 \pm 0.004$ & $0.761 \pm 0.004$ & $0.829 \pm 0.003$ & $0.402 \pm 0.007$ \\
% gemma3-4b & \textbf{0.788} $\pm 0.007$ & $0.189 \pm 0.002$ & $0.907 \pm 0.005$ & $0.761 \pm 0.005$ & $0.659 \pm 0.008$ \\
% mistral\_latest & $0.604 \pm 0.013$ & $0.144 \pm 0.003$ & $0.872 \pm 0.004$ & $0.755 \pm 0.010$ & $0.722 \pm 0.002$ \\
% llama3\_1\_8B & $0.537 \pm 0.034$ & $0.126 \pm 0.007$ & $0.853 \pm 0.006$ & $0.688 \pm 0.007$ & $0.418 \pm 0.001$ \\
% \hline
% ministral-3-14b & $0.439 \pm 0.009$ & $0.116 \pm 0.002$ & $0.673 \pm 0.006$ & $0.664 \pm 0.006$ & $0.675 \pm 0.023$ \\
% gemma3-27b & \textbf{0.737} $\pm 0.012$ & $0.170 \pm 0.003$ & $0.659 \pm 0.003$ & $0.584 \pm 0.004$ & $1.114 \pm 0.094$ \\
% \hline
% llama3\_70B & \textbf{0.766} $\pm 0.012$ & $0.183 \pm 0.002$ & $0.882 \pm 0.006$ & $0.737 \pm 0.009$ & $1.424 \pm 0.004$ \\
% \hline
% \end{tabular}
% }
% \caption{Experimental results on dataset $D_1$, highlighting the superior model per tier using recall as the primary evaluation metric.}
% \label{table: D1}
% \end{table*}

\begin{table*}[!h]
\centering
\resizebox{\textwidth}{!}{
\begin{tabular}{llccccc}
\hline
Tier & Model \cite{team2024gemma}, \cite{grattafiori2024llama} & Recall & Precision & Soft Recall & Soft Precision & Mean LLM time \\
\hline
\multirow{4}{*}{\rotatebox[origin=c]{0}{\textit{Low}}} & ministral-3-3b & $0.579 \pm 0.012$ & $0.214 \pm 0.004$ & $0.761 \pm 0.004$ & $0.829 \pm 0.003$ & $0.402 \pm 0.007$ \\
 & gemma3-4b & \textbf{0.788} $\pm 0.007$ & $0.189 \pm 0.002$ & $0.907 \pm 0.005$ & $0.761 \pm 0.005$ & $0.659 \pm 0.008$ \\
 & mistral\_latest & $0.604 \pm 0.013$ & $0.144 \pm 0.003$ & $0.872 \pm 0.004$ & $0.755 \pm 0.010$ & $0.722 \pm 0.002$ \\
 & llama3\_1\_8B & $0.537 \pm 0.034$ & $0.126 \pm 0.007$ & $0.853 \pm 0.006$ & $0.688 \pm 0.007$ & $0.418 \pm 0.001$ \\
\hline
\multirow{2}{*}{\rotatebox[origin=c]{0}{\textit{Medium}}} & ministral-3-14b & $0.439 \pm 0.009$ & $0.116 \pm 0.002$ & $0.673 \pm 0.006$ & $0.664 \pm 0.006$ & $0.675 \pm 0.023$ \\
 & gemma3-27b & \textbf{0.737} $\pm 0.012$ & $0.170 \pm 0.003$ & $0.659 \pm 0.003$ & $0.584 \pm 0.004$ & $1.114 \pm 0.094$ \\
\hline
\rotatebox[origin=c]{0}{\textit{High}} & llama3\_70B & \textbf{0.766} $\pm 0.012$ & $0.183 \pm 0.002$ & $0.882 \pm 0.006$ & $0.737 \pm 0.009$ & $1.424 \pm 0.004$ \\
\hline
\end{tabular}
}
\caption{Experimental results on dataset $D_1$, highlighting the superior model per tier using recall as the primary evaluation metric.}
\label{table: D1}
\end{table*}

\begin{table*}[!h]
\centering
\resizebox{\textwidth}{!}{
\begin{tabular}{llccccc}
\hline
Tier & Model \cite{team2024gemma}, \cite{grattafiori2024llama} & Recall & Precision & soft Recall & soft Precision & Mean LLM time \\
\hline
\multirow{4}{*}{\rotatebox[origin=c]{0}{\textit{Low}}} & ministral-3-3b & $0.640 \pm 0.012$ & $0.237 \pm 0.003$ & $0.765 \pm 0.006$ & $0.838 \pm 0.005$ & $0.394 \pm 0.001$ \\
& gemma3-4b & \textbf{0.801} $\pm 0.007$ & $0.192 \pm 0.001$ & $0.922 \pm 0.005$ & $0.797 \pm 0.007$ & $0.657 \pm 0.085$ \\
& mistral\_latest & $0.599 \pm 0.015$ & $0.145 \pm 0.003$ & $0.860 \pm 0.003$ & $0.743 \pm 0.004$ & $0.711 \pm 0.003$ \\
& llama3\_1\_8B & $0.528 \pm 0.013$ & $0.125 \pm 0.003$ & $0.845 \pm 0.005$ & $0.711 \pm 0.008$ & $0.418 \pm 0.002$ \\
\hline
\multirow{2}{*}{\rotatebox[origin=c]{0}{\textit{Medium}}} & ministral-3-14b & $0.423 \pm 0.010$ & $0.112 \pm 0.003$ & $0.641 \pm 0.004$ & $0.657 \pm 0.006$ & $0.654 \pm 0.003$ \\
& gemma3-27b & \textbf{0.708} $\pm 0.007$ & $0.166 \pm 0.002$ & $0.652 \pm 0.009$ & $0.585 \pm 0.011$ & $1.107 \pm 0.095$ \\
\hline
\rotatebox[origin=c]{0}{\textit{High}}  & llama3\_70B & \textbf{0.744} $\pm 0.017$ & $0.176 \pm 0.004$ & $0.890 \pm 0.008$ & $0.747 \pm 0.011$ & $1.440 \pm 0.003$ \\
\hline
\end{tabular}
}
\caption{Experimental results on dataset $D_2$, highlighting the superior model per tier using recall as the primary evaluation metric.}
\label{table: D2}
\end{table*}

\begin{table*}[!h]
\centering
\resizebox{\textwidth}{!}{
\begin{tabular}{llccccc}
\hline
Tier & Model \cite{team2024gemma}, \cite{grattafiori2024llama} & Recall & Precision & soft Recall & soft Precision & Mean LLM time \\
\hline
\multirow{4}{*}{\rotatebox[origin=c]{0}{\textit{Low}}} & ministral-3-3b & $0.430 \pm 0.003$ & $0.206 \pm 0.004$ & $0.572 \pm 0.009$ & $0.577 \pm 0.008$ & $0.278 \pm 0.003$ \\
& gemma3-4b & $0.572 \pm 0.003$ & $0.231 \pm 0.001$ & $0.685 \pm 0.012$ & $0.617 \pm 0.009$ & $0.313 \pm 0.004$ \\
& mistral\_latest & $0.530 \pm 0.023$ & $0.207 \pm 0.009$ & $0.631 \pm 0.011$ & $0.649 \pm 0.013$ & $0.322 \pm 0.006$ \\
& llama3\_1\_8B & \textbf{0.631} $\pm 0.016$ & $0.251 \pm 0.004$ & $0.759 \pm 0.009$ & $0.650 \pm 0.008$ & $0.300 \pm 0.008$ \\
\hline
\multirow{2}{*}{\rotatebox[origin=c]{0}{\textit{Medium}}} & ministral-3-14b & $0.454 \pm 0.003$ & $0.196 \pm 0.001$ & $0.585 \pm 0.003$ & $0.551 \pm 0.002$ & $0.323 \pm 0.009$ \\
& gemma3-27b & \textbf{0.683} $\pm 0.016$ & $0.258 \pm 0.005$ & $0.707 \pm 0.012$ & $0.618 \pm 0.009$ & $0.417 \pm 0.015$ \\
\hline
\rotatebox[origin=c]{0}{\textit{Low}} & llama3-70B & \textbf{0.622} $\pm 0.010$ & $0.274 \pm 0.007$ & $0.789 \pm 0.006$ & $0.793 \pm 0.006$ & $0.417 \pm 0.002$ \\
\hline
\end{tabular}
}
\caption{Experimental results on dataset $D_3$, highlighting the superior model per tier using recall as the primary evaluation metric.}
\label{table: D3}
\end{table*}

\begin{lstlisting}[caption={Single dataset entry of the KG enhanced by Gemma3:4B.}, label={lst:gemma}]
PREFIX MLtask:  <https://secured-project.eu/MLtask/>
PREFIX dataset: <https://secured-project.eu/dataset/>
PREFIX ex:      <https://secured-project.eu/>
PREFIX person:  <https://secured-project.eu/person/>
PREFIX tag:     <https://secured-project.eu/tag/>
PREFIX xsd:     <http://www.w3.org/2001/XMLSchema#>

dataset:yasserh_breast-cancer-dataset
        a                  ex:Dataset;
        ex:contentSize     0.049794e0;
        ex:downloads       111270;
        ex:hasDescription  "Binary Classification Prediction for type of Breast Cancer";
        ex:hasName         "Breast Cancer Dataset";
        ex:isRelatedTo     tag:cancer , tag:classification , tag:binary_classification , tag:healthcare , tag:tabular;
        (*@\textcolor{highlight}{\textbf{ex:isSuitableFor   MLtask:classification , MLtask:binary\_classification , MLtask:logistic\_regression;}}@*)
        ex:ownedBy         person:m-yasser-h;
        ex:publishedDate   "2021-12-29 19:07:20.320";
        ex:upvoteCount     649;
        ex:usabilityScore  1.0e0 .

tag:binary_classification
        a       ex:Tag .

tag:healthcare  a  ex:Tag .

tag:tabular  a  ex:Tag .

person:m-yasser-h  a  ex:Person;
        ex:hasName  "M Yasser H" .
        
\end{lstlisting}

\begin{figure}[!h]
    \centering
    \includegraphics[width=0.9\linewidth]{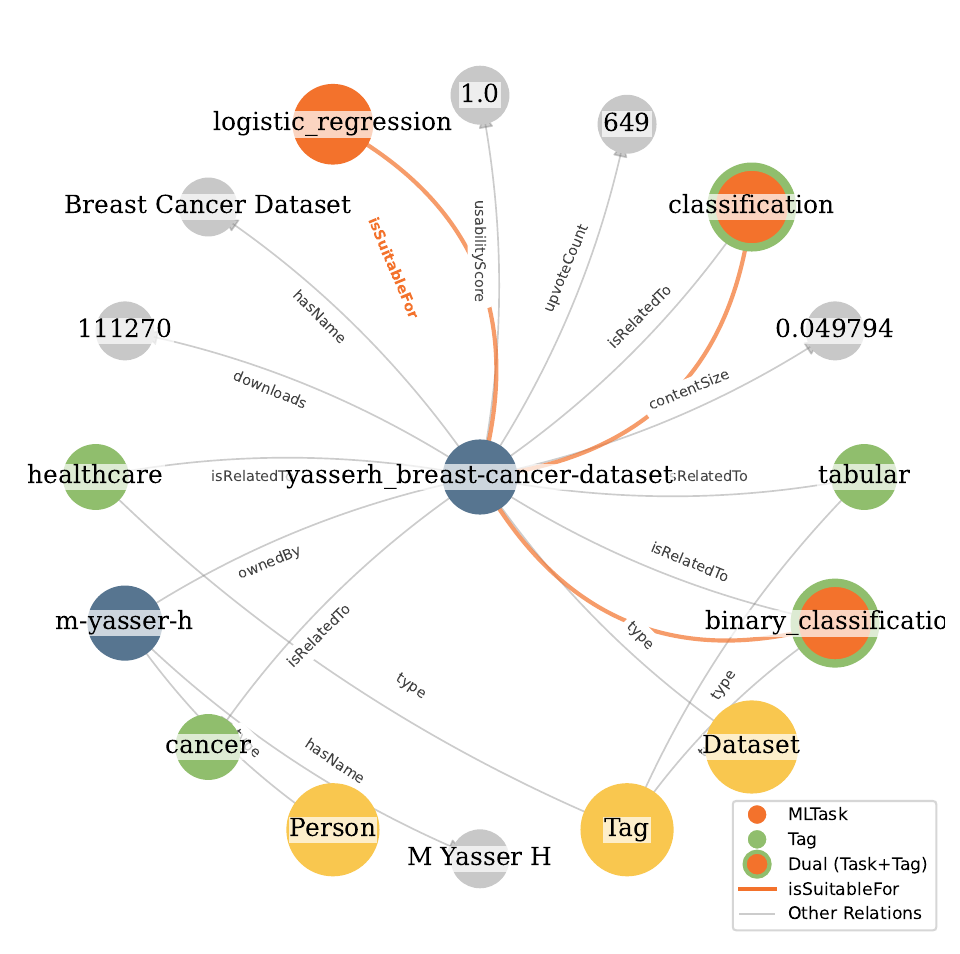}
    \caption{Graphical representation of a single KG entry enhanced by Gemma3:4B's metadata annotation tags.}
    \label{fig:single_dataset_gemma}
\end{figure}

\begin{lstlisting}[caption={Single dataset entry of the KG enhanced by Llama3.1:70B.}, label={lst:llama}]
PREFIX MLtask:  <https://secured-project.eu/MLtask/>
PREFIX dataset: <https://secured-project.eu/dataset/>
PREFIX ex:      <https://secured-project.eu/>
PREFIX person:  <https://secured-project.eu/person/>
PREFIX tag:     <https://secured-project.eu/tag/>
PREFIX xsd:     <http://www.w3.org/2001/XMLSchema#>

dataset:yasserh_breast-cancer-dataset
        a                  ex:Dataset;
        ex:contentSize     0.049794e0;
        ex:downloads       111270;
        ex:hasDescription  "Binary Classification Prediction for type of Breast Cancer";
        ex:hasName         "Breast Cancer Dataset";
        ex:isRelatedTo     tag:cancer , tag:classification , tag:binary_classification , tag:healthcare , tag:tabular;
        (*@\textcolor{highlight}{\textbf{ex:isSuitableFor   MLtask:classification , MLtask:binary\_classification;}}@*)
        ex:ownedBy         person:m-yasser-h;
        ex:publishedDate   "2021-12-29 19:07:20.320";
        ex:upvoteCount     649;
        ex:usabilityScore  1.0e0 .

tag:binary_classification
        a       ex:Tag .

tag:healthcare  a  ex:Tag .

tag:tabular  a  ex:Tag .

person:m-yasser-h  a  ex:Person;
        ex:hasName  "M Yasser H" .
        
\end{lstlisting}

\begin{figure}[!h]
    \centering
    \includegraphics[width=0.9\linewidth]{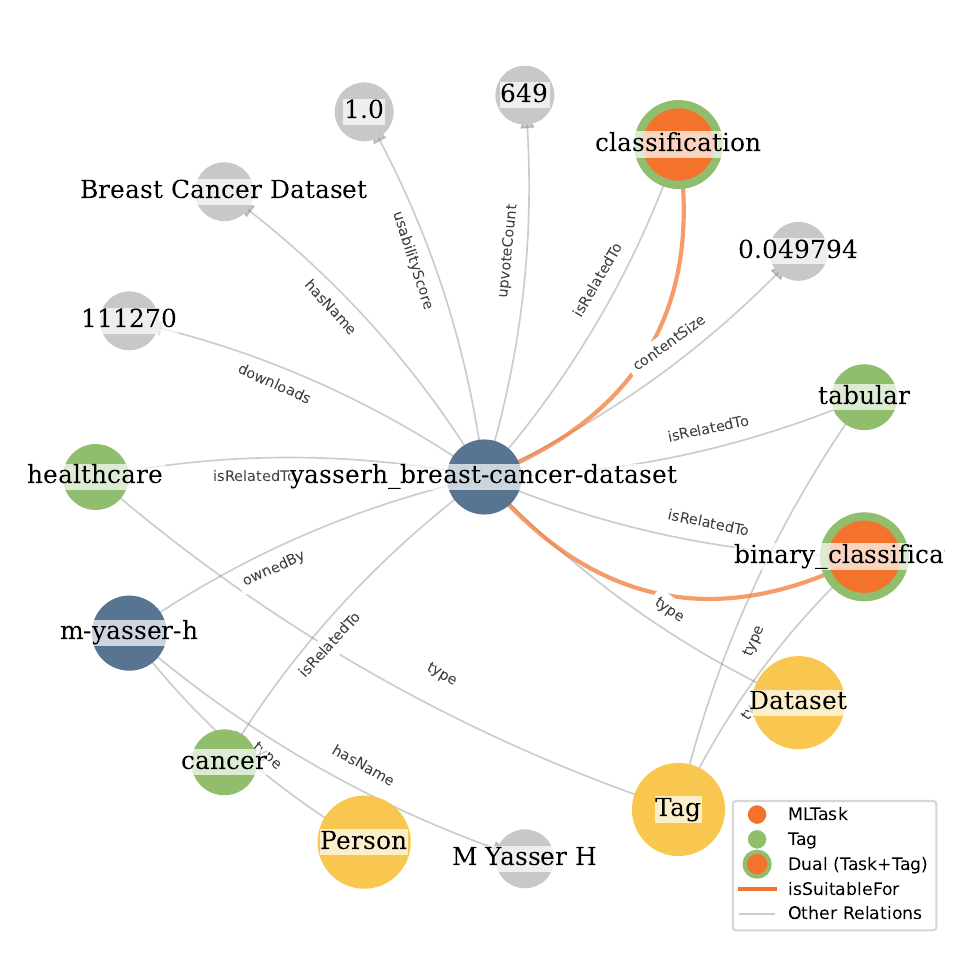}
    \caption{Graphical representation of a single KG entry enhanced by Llama3.1:70B's metadata annotation tags.}
    \label{fig:single_dataset_llama}
\end{figure}

\subsection{Knowledge Graph visualization}

To qualitatively examine the model outputs, we present a single entry from $D_1$ as enhanced by Gemma3:4B and Llama3.1:70B in Listings \ref{lst:gemma} and \ref{lst:llama}, respectively. In both listings, the new relation 'isSuitableFor', along with the identified ML tasks, has been highlighted with orange-colored font. The corresponding graphical representations are shown in Figures \ref{fig:single_dataset_gemma} and \ref{fig:single_dataset_llama}. Yellow nodes depict the main ontology classes, such as Person, Tag, and Dataset. Blue components represent instances of these classes, while gray elements show the actual values of the entry’s fields. Green nodes represent user-provided tags, and orange nodes represent the LLM-identified ML tasks. In most cases, a non-empty set of nodes occurs at the intersection of tags and ML tasks. These nodes are colored orange and highlighted with a green border. As observed, both models have classified the tags 'classification' and 'binary\_classification' as suitable for ML tasks, while Gemma3:4B proposes the additional 'logistic regression' keyword as ML task. 

\begin{figure*}[!t]
    \centering
    \includegraphics[width=0.96\linewidth]{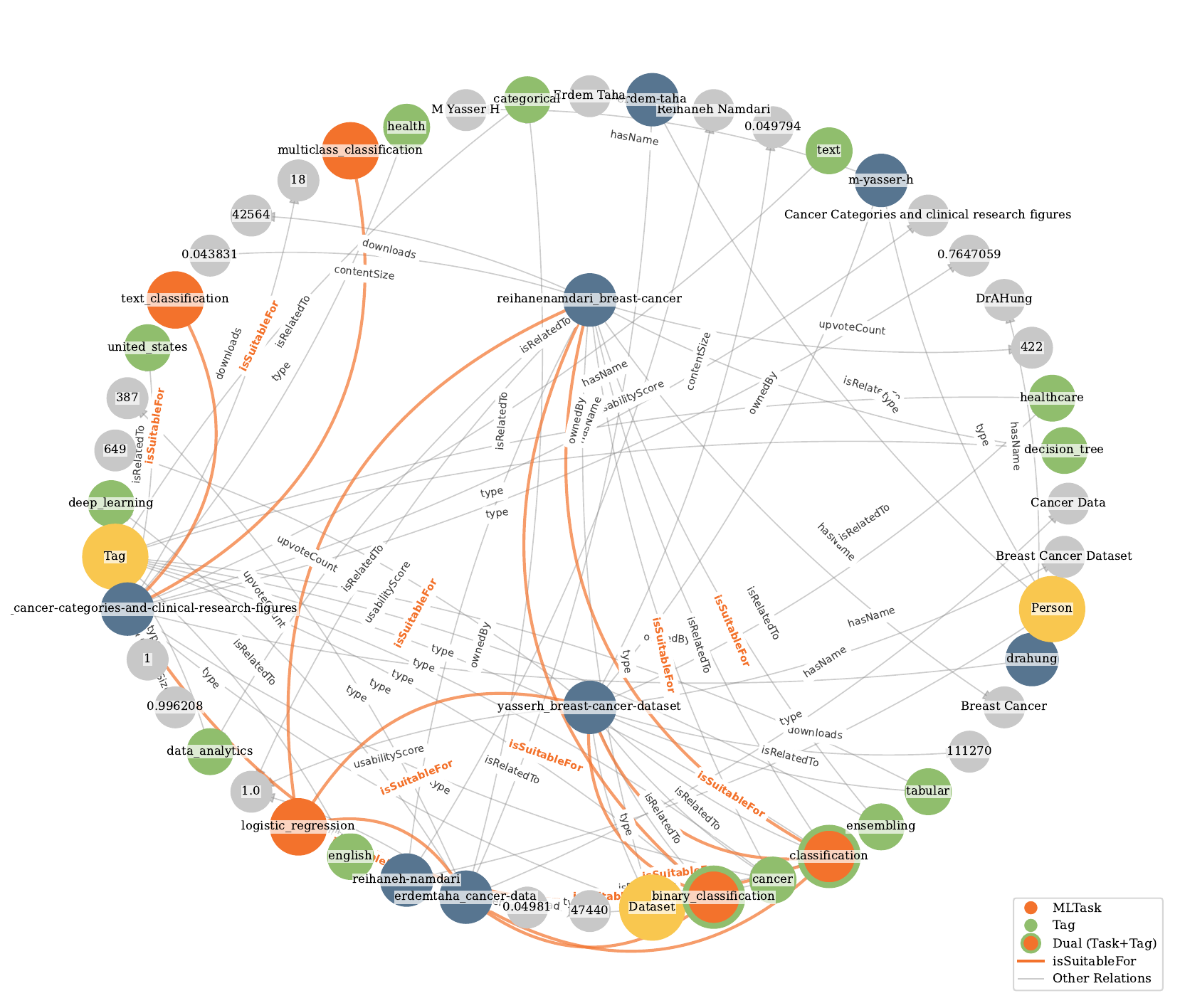}
    \caption{A tiny subset of the Knowledge Graph enhanced by the Gemma3-4B's metadata annotation tags. }
    \label{fig:kg_gemma}
\end{figure*}

In Figure \ref{fig:kg_gemma}, we present a subset of the final KG augmented by the Gemma 3-4B model, which illustrates information from the four datasets in $D_1$. The new connections are shown as orange-colored edges originating from the dataset nodes and leading to the corresponding ML tasks, signifying the 'isRelatedTo' relation. Additionally, two existing tags (green nodes) and three newly added nodes have been identified as ML tasks.

% \begin{table*}[!h]
% \centering
% \resizebox{\textwidth}{!}{
% \begin{tabular}{lccccc}
% \hline
% Model & Recall & Precision & soft Recall & soft Precision & Mean LLM time \\
% \hline
% ministral-3-3b & $0.640 \pm 0.012$ & $0.237 \pm 0.003$ & $0.765 \pm 0.006$ & $0.838 \pm 0.005$ & $0.394 \pm 0.001$ \\
% gemma3-4b & $0.801 \pm 0.007$ & $0.192 \pm 0.001$ & $0.922 \pm 0.005$ & $0.797 \pm 0.007$ & $0.657 \pm 0.085$ \\
% mistral\_latest & $0.599 \pm 0.015$ & $0.145 \pm 0.003$ & $0.860 \pm 0.003$ & $0.743 \pm 0.004$ & $0.711 \pm 0.003$ \\
% llama3\_1\_8B & $0.528 \pm 0.013$ & $0.125 \pm 0.003$ & $0.845 \pm 0.005$ & $0.711 \pm 0.008$ & $0.418 \pm 0.002$ \\
% \hline
% ministral-3-14b & $0.423 \pm 0.010$ & $0.112 \pm 0.003$ & $0.641 \pm 0.004$ & $0.657 \pm 0.006$ & $0.654 \pm 0.003$ \\
% gemma3-27b & $0.708 \pm 0.007$ & $0.166 \pm 0.002$ & $0.652 \pm 0.009$ & $0.585 \pm 0.011$ & $1.107 \pm 0.095$ \\
% \hline
% llama3\_70B & $0.744 \pm 0.017$ & $0.176 \pm 0.004$ & $0.890 \pm 0.008$ & $0.747 \pm 0.011$ & $1.440 \pm 0.003$ \\
% \hline
% \end{tabular}
% }
% \caption{Results on cancer data}
% \label{table: D4}
% \end{table*}

% \begin{figure*}[!t]
%     \centering
%     \includegraphics[width=1\linewidth]{figs/graphs/llama_cancer.pdf}
%     \caption{A tiny subset of the Knowledge Graph enhanced by the Lamma3-70B LLM}
%     \label{fig:kg_llama}
% \end{figure*}

%% file: sections/conclusion.tex
In this work, we present a data lake architecture designed according to modern standards and state-of-the-art mechanisms to organize data for integrity, reachability, scalability, and adaptability. The proposed system adopts a hybrid approach; it employs a document-based solution to store essential metadata, thereby building a rigid library for indexing. Simultaneously, semantic triples are generated to dynamically construct a knowledge graph, providing a schema-on-read layer that promotes flexibility in extracting knowledge from raw metadata.

On top of this, we propose a hybrid construction pipeline for the Knowledge Graph. This approach combines rule-based RDF triple generation with LLM-driven knowledge synthesis, where the LLM is prompted as an AI expert in dataset and model categorization. This ensures a foundation of trustworthy entries derived from rigid metadata rules, which are then enhanced by associative insights from the LLM. While this promotes flexibility in knowledge extraction, it risks generating redundant or 'noisy' metadata. To mitigate this, we incorporate a human-in-the-loop verification step to ensure the essentiality and relevance of all LLM-generated content.

We established three distinct metadata collections derived from public datasets and ML models hosted on Kaggle. To evaluate semantic extraction, we selected seven Large Language Models (LLMs) from three distinct architectural families, categorizing them into three tiers based on parameter cardinality. Each model was prompted to act as an AI specialist in dataset and model categorization, tasked with discovering essential tags for the extraction of rich knowledge expressed as RDF triples.Experimental results indicate that Gemma 3:4B outperformed the other models in most scenarios, with the notable exception of the $D_3$ case, where Llama 3.1:8B demonstrated superior performance. Based on these results, we selected Gemma 3:4B for our case study. This selection does not imply that the remaining models are unsuitable for such tasks; on the contrary, all tested models performed well. Our choice was primarily an architectural decision, prioritizing a high performance-to-latency ratio. For alternative tasks, the selection criteria may be adapted accordingly. Future work will involve designing a scheduled LLM intervention to maintain the Knowledge Graph (KG) by updating or adding new tags. In such scenarios, a medium- or high-tier model may prove more suitable to handle the increased complexity.

It should be noted that the proposed system can be used in various applications, including remote sensing, as shown in the \href{https://dawetrest.eu/}{DaWeRest} project. The medical domain was selected as an illustrative use case for this work.

\section*{Acknowledgement}
This paper is funded by the EU funded project DaWeRest “Danube Wetlands and flood plains Restoration through systemic, community engaged and sustainable innovative actions” with grant agreement no. 101113015.

The authors used Gemini and ChatGPT to check grammar and improve the readability of the manuscript. The authors reviewed and edited all output and take full responsibility for the content of the article.

\clearpage

\begin{figure*}[!t]
    \centering
    \includegraphics[width=1\linewidth]{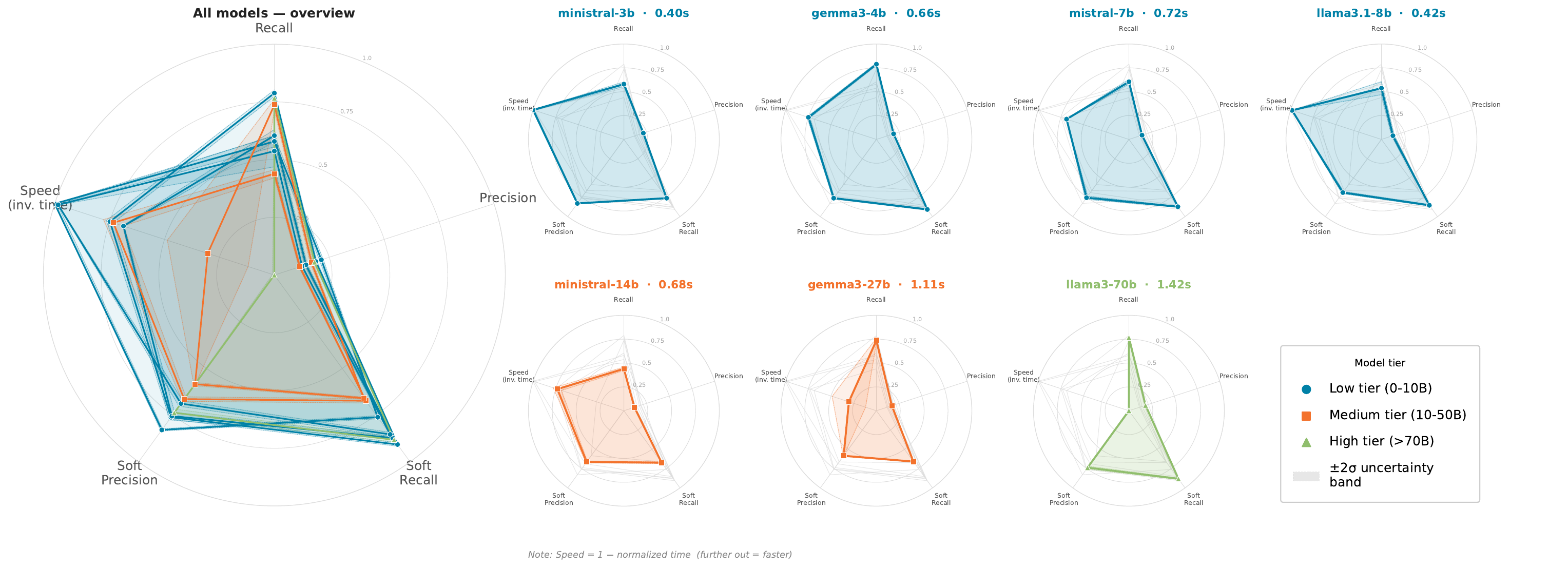}
    \caption{Performance of the selected models on dataset $D_1$. The large plot on the left depicts the metrics of all models, while each of the small plots corresponds to the performance of a single model. In this case, $Gemma3:4b$ appears to be the superior model.}
    \label{fig:d1_results}
\end{figure*}
\begin{figure*}[!h]
    \centering
    \includegraphics[width=1\linewidth]{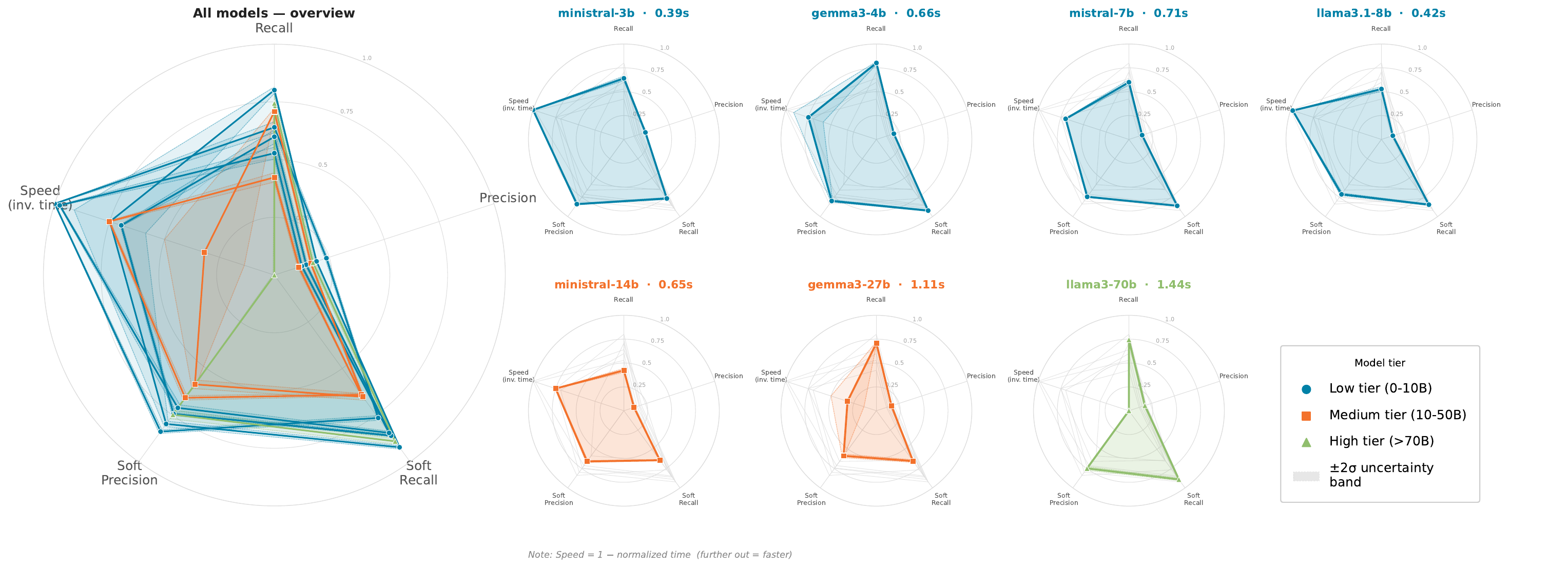}
    \caption{Performance of the selected models on dataset $D_2$. The large plot on the left depicts the metrics of all models, while each of the small plots corresponds to the performance of a single model. In this case, $Gemma3:4b$ appears to be the superior model.}
    \label{fig:d2_results}
\end{figure*}
\begin{figure*}[!h]
    \centering
    \includegraphics[width=1\linewidth]{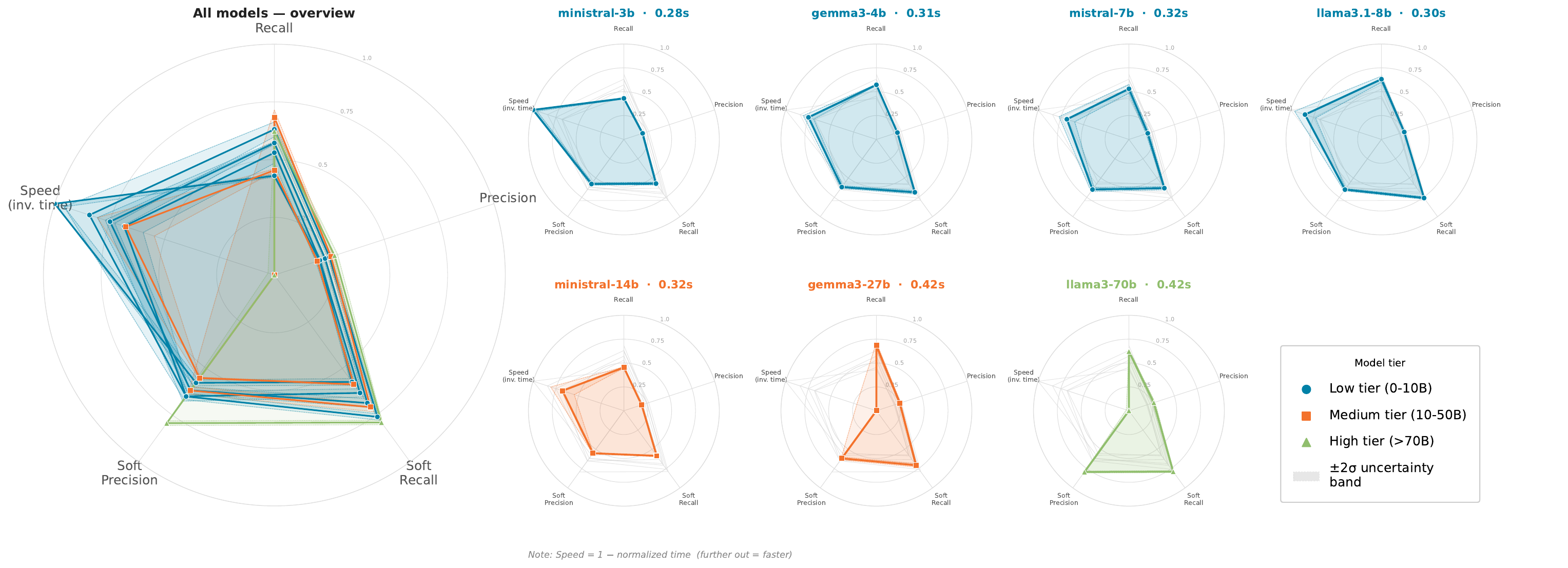}
    \caption{Performance of the selected models on dataset $D_3$. The large plot on the left depicts the metrics of all models, while each of the small plots corresponds to the performance of a single model. In this case, $Llama3.1:8b$ appears to be the superior model.}
    \label{fig:d3_results}
\end{figure*}